\documentclass[english]{article}
\usepackage{geometry}
\usepackage[T1]{fontenc}
\usepackage{microtype}
\usepackage{graphicx}
\usepackage{booktabs}
\usepackage{amsmath}
\usepackage{amssymb}
\usepackage{algorithm}
\usepackage{algorithmic}
\usepackage{mathtools}
\usepackage{amsthm}
\usepackage{bbding}
\usepackage{multirow}
\usepackage{xspace}
\usepackage{hyperref}
\usepackage[capitalize,noabbrev]{cleveref}
\usepackage{natbib}
\renewcommand{\cite}{\citep}
\def\eqref#1{equation~\ref{#1}}

\usepackage{titletoc}
\usepackage{xspace}
\newcommand{\name}{\texttt{SCALAR}\xspace}

\usepackage{amsmath,amssymb,amsthm,bm}

\def\eqref#1{equation~\ref{#1}}

\def\1{\bm{1}}

\DeclareMathAlphabet{\mathsfit}{\encodingdefault}{\sfdefault}{m}{sl}
\SetMathAlphabet{\mathsfit}{bold}{\encodingdefault}{\sfdefault}{bx}{n}

\theoremstyle{plain}
\newtheorem{theorem}{Theorem}[section]

\theoremstyle{definition}

\newtheorem{assumption}[theorem]{Assumption}
\theoremstyle{remark}

\usepackage{hyperref}
\usepackage{url}

\usepackage[T1]{fontenc}
\usepackage{algorithmic}
\usepackage{algorithm}
\usepackage{tcolorbox}
\usepackage{graphicx}
\usepackage{bbding}
\usepackage{subcaption}
\usepackage{booktabs}
\usepackage{multirow}
\usepackage{titletoc}
\renewcommand{\cite}{\citep}
\usepackage{color, colortbl}
\definecolor{blue1}{rgb}{0.6, 0.6, 0.6}
\definecolor{blue2}{rgb}{0.75, 0.8, 0.99}
\definecolor{blue3}{rgb}{0.84,0.89,0.99}
\definecolor{blue4}{rgb}{0.93,0.95,0.99}

\definecolor{DarkPink}{rgb}{0.5,0.0,0.18}
\definecolor{DarkGreen}{rgb}{0.1,0.5,0.1}
\definecolor{DarkRed}{rgb}{0.5,0.1,0.1}
\definecolor{DarkBlue}{rgb}{0.1,0.1,0.7}
\definecolor{DarkYellow}{rgb}{.79,.79,0}
\hypersetup{
    unicode=false,          
    pdftoolbar=true,        
    pdfmenubar=true,        
    pdffitwindow=false,      
    pdfnewwindow=true,      
    colorlinks=true,       
    linkcolor=DarkBlue,          
    citecolor=DarkRed,        
    filecolor=DarkRed,      
    urlcolor=DarkBlue,          
    pdftitle={},
    pdfauthor={},
}

\begin{document}

\title{Direction or Magnitude? Learning Scalars over Task-Calibrated Directions}
\author{
  Ziqing Wen$^{1,*}$,
  Xiaohui Song$^{2}$,
  Jiahuan Wang$^{1}$,
  Ping Luo$^{1}$,
  Junlin Zeng$^{1}$, \\
  Dongsheng Li$^{1}$,
  Kele Xu$^{1}$,
  Tao Sun$^{1,\dag}$
}
\date{}

\maketitle

\renewcommand*{\thefootnote}{\fnsymbol{footnote}}
\footnotetext[1]{Work done during an internship at OPPO AI-Center.}

\renewcommand*{\thefootnote}{\arabic{footnote}}

\footnotetext[1]{National University of Defense Technology. Contact: \texttt{suntao.saltfish@outlook.com}}
\footnotetext[2]{OPPO AI center}

\begin{abstract}
Parameter-efficient fine-tuning (PEFT) aims to preserve adaptation
quality while reducing training costs. We propose \textbf{\name},
which achieves performance comparable to LoRA with substantially
fewer trainable parameters, lower peak training memory, and higher
throughput. \name constructs a fixed input basis from task activations
and assigns a random output dictionary to each input direction.
Fine-tuning learns only layer-specific scalar coefficients, reducing
gradient and optimizer-state storage. Grouped computation aggregates
output atoms before applying the adapter to token activations, while
dictionary sharing across compatible layers reduces fixed-basis
storage. Experiments on commonsense reasoning, mathematical reasoning,
code generation, and knowledge evaluation after instruction tuning
demonstrate effective adaptation at small trainable budgets.
On commonsense reasoning, \name achieves accuracy comparable to LoRA with about 4\% of its trainable parameters.
System measurements on Qwen2.5-7B show 20\% lower peak training memory
and 6\% higher throughput than LoRA. The combination of adaptation
quality, a compact trainable state, and training efficiency makes
\name a promising approach to resource-constrained fine-tuning.
\end{abstract}

\section{Introduction}

Large language models (LLMs) achieve strong general-purpose performance,
but full-parameter fine-tuning incurs substantial training memory and
optimization costs. Parameter-efficient fine-tuning (PEFT) alleviates
these costs by updating a small subset of parameters, with
LoRA~\cite{hu2022lora} serving as a widely adopted baseline. For an
$m\times n$ linear layer, LoRA optimizes $R(m+n)$ parameters together
with their gradients and optimizer states. These costs are especially
relevant to on-device personalization, where local model updates must
operate within limited memory and compute budgets. Retaining adaptation
quality with a smaller trainable state, lower peak memory, and higher
throughput can make such local updates more practical.

Recent methods further reduce the optimization budget by learning compact
coefficients over fixed bases
~\cite{balazy2024loraxs,kopiczko2024vera,abbasi2024nola,
albert2025randlora}. Their effectiveness depends on both the choice of
update directions and the computation required to apply them. Random
bases provide inexpensive candidate directions, while weight- or
activation-derived subspaces use information from pretrained models or
downstream data
~\cite{meng2024pissa,yang2024corda,paischer2026parametereva}.
Recent evidence also suggests that useful task-improving perturbations
can be abundant around pretrained weights~\cite{gan2026neural}. These
observations motivate the following question.

\begin{tcolorbox}[
    colback=pink!30,
    colframe=pink!60!black,
    boxrule=0.8pt,
    arc=6pt,
    left=6pt, right=6pt, top=6pt, bottom=6pt
]
\begin{center}
\textit{How can we construct a task-relevant fixed update space that
maintains adaptation quality with few trainable coefficients, low
memory use, and high throughput?}
\end{center}
\end{tcolorbox}

We first investigate this question through a learned-direction reweighting
diagnostic. We train a rank-one LoRA adapter, freeze its learned update
direction for each adapted matrix, and re-optimize one scalar coefficient
per matrix. On a 7B model, the reweighting stage optimizes $196$
parameters compared with LoRA's $2{,}523{,}136$, and achieves similar
optimization behavior and downstream accuracy. This observation motivates
constructing informative directions before fine-tuning and adapting
through a small set of coefficients.

We propose \textbf{\name: Subspace-Calibrated Adaptation via Layerwise
Atom Reweighting}. \name constructs a fixed input basis from the leading
singular vectors of downstream activations and assigns a separate random
output dictionary to each input direction. Learning the coefficients
selects an effective output vector within each of these fixed output
subspaces. With $R$ input directions and $H$ output atoms per direction,
each adapted matrix has $K=RH$ trainable coefficients and an update rank
of at most $R$. The output atoms in each group are aggregated before
application to the activations, so the token-dependent matrix
multiplications operate on $R$ effective directions. Compatible layers
can also share the frozen output dictionaries.

Experiments on commonsense reasoning, mathematical reasoning, and code
generation show competitive performance with compact trainable budgets.
On Qwen2.5-7B and LLaMA3-8B, \name approaches rank-8 LoRA accuracy on
commonsense reasoning using about 4\% of its trainable parameters.
System measurements demonstrate lower peak training memory and higher
throughput, giving the compact parameterization practical value during
training. Our analysis builds on activation coverage to examine the
initial gradient signal and connects it to the coefficients trained by
\name.

\begin{figure*}[!t]
    \centering
    \begin{subfigure}[b]{0.45\linewidth}
        \centering
        \includegraphics[width=\linewidth]{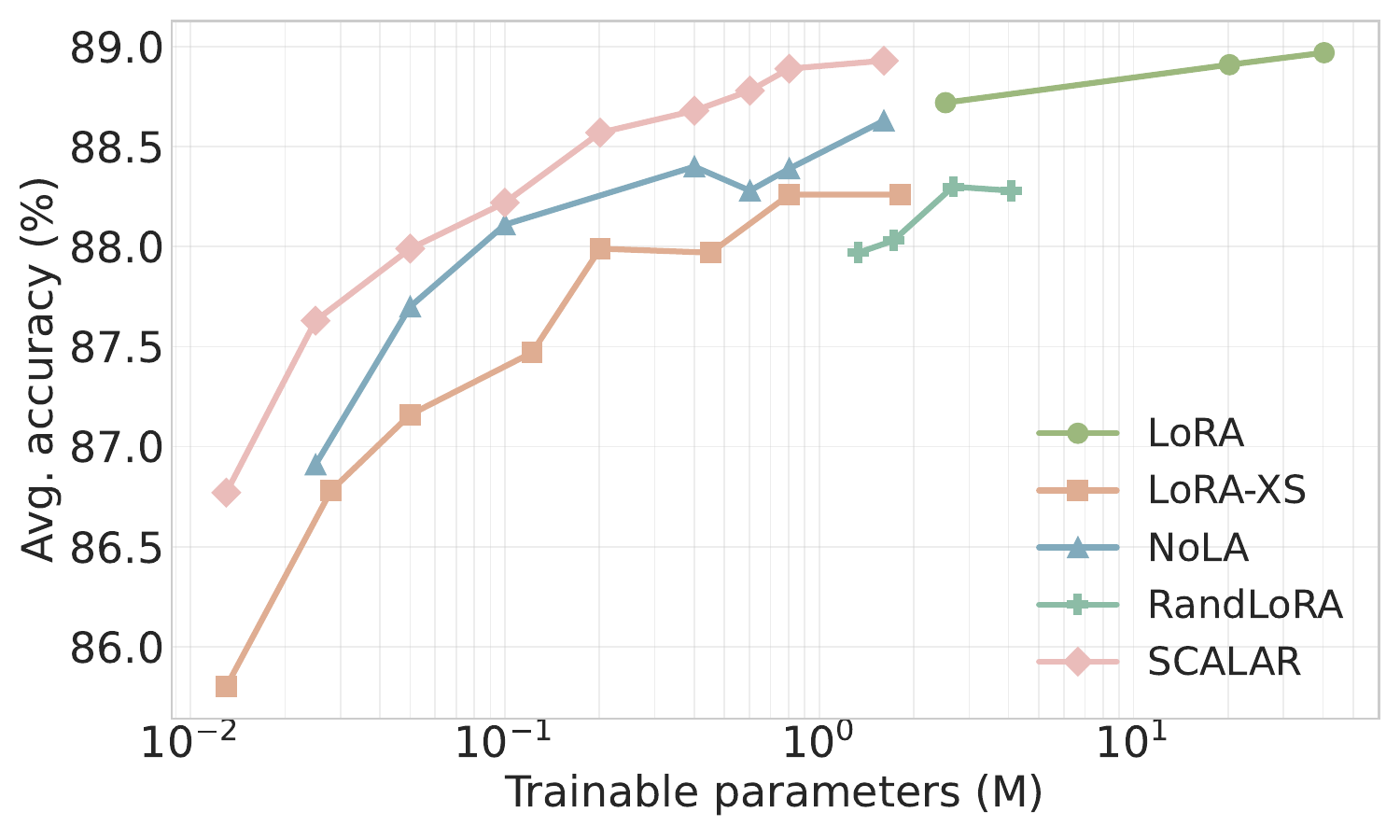}
    \caption{Qwen2.5-7B}
    \label{fig:common_7b_params}
    \end{subfigure}
    \hfill
    \begin{subfigure}[b]{0.45\linewidth}
        \centering
        \includegraphics[width=\linewidth]{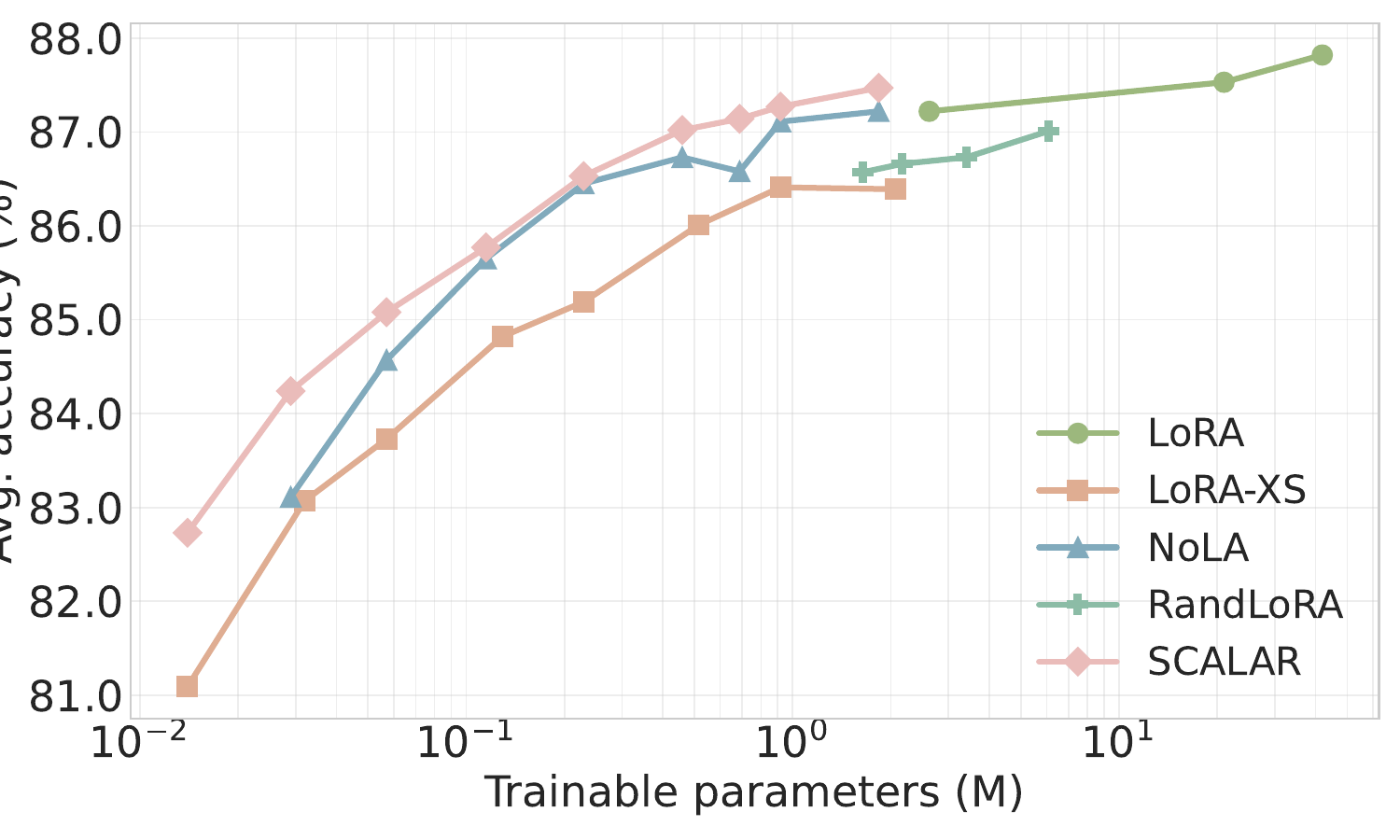}
        \caption{LLaMA3-8B}
        \label{fig:common_8b_params}
    \end{subfigure}
    \caption{\textbf{Average accuracy on 8 commonsense reasoning tasks across trainable parameter budgets.} \name approaches LoRA-level accuracy with substantially fewer trainable parameters.}
    \label{fig:common_param}
\end{figure*}

Our main contributions are summarized as follows.

\noindent\textbf{Task-calibrated fixed-subspace adaptation.}
We introduce \name, which combines an activation-derived input basis
with a separate random output subspace for each input direction. Its
grouped parameterization learns $RH$ coefficients and applies the update
through $R$ aggregated output vectors. Compatible layers share their
fixed output dictionaries to reduce basis storage.

\noindent\textbf{Analysis and diagnostic evidence.}
Building on activation-subspace analysis, we characterize the effect of
gradient-weighted activation geometry on the initial learning signal and
derive its connection to \name's coefficients. A learned-direction
reweighting diagnostic motivates adaptation within informative fixed
update spaces.

\noindent\textbf{Competitive performance and training efficiency.}
\name achieves competitive adaptation performance with substantially
fewer trainable parameters across commonsense reasoning, mathematical
reasoning, code generation, and general-domain knowledge evaluation.
System measurements on Qwen2.5-7B show lower peak training memory and
higher throughput than LoRA, NOLA, and RandLoRA. Structural ablations
examine the benefits of grouped computation and shared output
dictionaries. The experiments cover three model families and multiple
parameter scales up to 32B, with additional large-model evaluations in
the Appendix.

\section{Related Work}

\begin{table*}[!th]
    \centering
    \caption{\textbf{Comparison of trainable parameters across methods.} Let $\bm{W}\in\mathbb{R}^{m\times n}$ and $d=\min(m,n)$. LoRA, LoRA-XS, VeRA, NOLA, and \name use rank $R$. NOLA uses $k_A$ and $k_B$ random bases, while \name uses $H$ random output atoms per input direction.}
    
    \resizebox{\linewidth}{!}{
    {\begin{tabular}{l|cccccc}
    \toprule
    & LoRA & LoRA-XS & VeRA & NOLA & RandLoRA & \cellcolor{blue4}\name \\
      \midrule
      $ \Delta \bm{W}$ & $\bm{B}\bm{A}$ & $\bm{B}\bm{M}\bm{A}$ & $\bm{\Lambda}_{b} \bm{B} \bm{\Lambda}_{d} \bm{A}$ & $\left(\sum_{i=1}^{k_B}\alpha_i\bm{B}_{i}\right)\left(\sum_{j=1}^{k_A}\beta_j\bm{A}_{j}\right)$ & $\sum_{j=1}^{q} \bm{B}_j \bm{\Lambda}_j \bm{A} \bm{\Gamma}_j$ & \cellcolor{blue4}$\sum_{r=1}^{R}\sum_{h=1}^{H}\gamma_{r,h}\bm{b}_{r,h}\bm{a}_{r}^{\top}$ \\
      Trainable params & $R(m+n)$ & $R^{2}$ & $m+R$ & $k_A+k_B$ & $\lceil d/R\rceil(d+R)$ & \cellcolor{blue4}$RH$ \\
    \bottomrule
    \end{tabular}}
    }
    \label{tab:param_comparison_baseline}
\end{table*}

LoRA~\cite{hu2022lora} adapts pretrained weights through two trainable
low-rank factors. Subsequent methods improve rank allocation,
parameterization, optimization, and routing
~\cite{zhang2023adalora,zhang2024autolora,liu2024dora,
tian2024hydralora,wang2024loraga,wang2025milora,
guo2025nlora,morris2026learning13}. Initialization also shapes the
adaptation subspace. PiSSA~\cite{meng2024pissa} uses pretrained weight
directions, EVA~\cite{paischer2026parametereva} uses dominant input
activation directions, and CorDA~\cite{yang2024corda} combines weights
with task-dependent activation statistics. Astra~\cite{liu2026astra}
initializes trainable LoRA factors from the tail eigenspaces of output
activations.

LoRA-FA~\cite{zhang2023lorafa} freezes a randomly initialized input factor
$\bm{A}$ and trains only $\bm{B}$, reducing adapter activation storage.
Zhu et al.~\cite{zhu2024asymmetry} study the distinct roles of the two
factors and the effectiveness of training only the output factor.
TLoRA~\cite{lin2026tlora} constructs $\bm{A}$ from pretrained weights
and input activation statistics, then freezes it while training
$\bm{B}$.

Fixed-basis methods further compress the trainable state.
LoRA-XS~\cite{balazy2024loraxs} learns a small core between fixed
SVD-derived projections, while NoRA~\cite{li2025nora} freezes outer
factors initialized by activation-aware SVD and trains a nested low-rank
structure. SVFT~\cite{lingam2024svft} learns sparse coefficients over
outer products of weight singular vectors. Other methods use random,
frequency-domain, or tensor bases
~\cite{nooralinejad2023pranc,gao2024parameter_fourier,gu2026tera}.
NOLA~\cite{abbasi2024nola} learns combinations of random matrices,
VeRA~\cite{kopiczko2024vera} scales shared random projections, and
RandLoRA~\cite{albert2025randlora} combines random low-rank bases with
diagonal scalings. \name pairs activation-derived input directions with
separate fixed random output dictionaries. Its coefficients determine
an effective output vector for each input direction, supporting compact
adaptation and grouped computation.

\section{Motivation}
\label{sec:motivation}

We motivate \name through a learned-direction reweighting diagnostic and
an analysis of input subspaces. The diagnostic examines adaptation through
scalar coefficients after informative directions have been obtained.
The analysis then connects dominant activation directions to input
coverage and the initial gradient signal, providing the basis for
constructing a fixed input subspace before fine-tuning.

\subsection{Learned-Direction Reweighting}
\label{sec:motivation_direction_reweighting}

We begin with a diagnostic experiment that separates adaptation directions
from their magnitudes. Any trained rank-$r$ LoRA update can be decomposed
through singular value decomposition as
\begin{equation}
\Delta\bm{W}_{\mathrm{LoRA}}
=
\sum_{i=1}^{r}
\sigma_i
\bm{u}_i
\bm{v}_i^{\top},
\nonumber
\end{equation}
where each pair $(\bm{u}_i,\bm{v}_i)$ defines a rank-one adaptation
direction and $\sigma_i$ specifies its magnitude. We freeze these
directions and re-optimize only their scalar coefficients:
\begin{equation}
\Delta\bm{W}
=
\sum_{i=1}^{r}
\gamma_i
\bm{u}_i
\bm{v}_i^{\top}.
\nonumber
\end{equation}

We report the Qwen2.5-7B results in
Table~\ref{tab:learned_direction_reweighting}. In this setting, a rank-one
LoRA adapter contains 2{,}523{,}136 trainable parameters, whereas the
learned-direction reweighting model optimizes only 196 coefficients.
Despite this substantial reduction, it closely follows the optimization
behavior of LoRA and achieves slightly higher average accuracy. The same
trend is observed on LLaMA3-8B; detailed training trajectories and
additional results are provided in
Appendix~\ref{app:oracle_directions}.

\begin{table}[!ht]
\centering
\caption{\textbf{Accuracy of LoRA and learned-direction reweighting on
8 commonsense reasoning tasks.} Learned-direction reweighting
achieves comparable performance while optimizing fewer than 0.01\%
as many parameters as LoRA.}
\label{tab:learned_direction_reweighting}
\resizebox{.95\linewidth}{!}{
\begin{tabular}{l|l|cccccccc|c}
\toprule
Method & Params & BoolQ & PIQA & SIQA & HS & WG & ARC-E & ARC-C & OBQA & Avg. \\
\midrule
LoRA & 2,523,136 & 74.1 & 90.3 & 81.8 & 95.7 & 88.7 & 96.2 & 89.4 & 91.8 & 88.5 \\
Learned-direction reweighting
& 196 & 73.9 & 90.1 & 81.8 & 96.0 & 89.0 & 96.4 & 89.9 & 92.4 & 88.7 \\
\bottomrule
\end{tabular}
}
\end{table}

These observations show that reweighting informative learned directions
can recover similar fitting behavior and downstream accuracy. They
motivate the practical design of \name, which constructs its input
directions from activation statistics before fine-tuning and learns
coefficients within the resulting fixed update space.

\subsection{Theoretical Analysis}

We analyze the role of the input subspace through activation coverage and
the initial gradient energy of a low-rank output factor. The analysis
characterizes the effect of weighting activations by the task gradient
and motivates the fixed input directions used by \name.

\subsubsection{Input-Side Subspace in Low-Rank Adaptation}

Consider a frozen linear transformation with a LoRA branch:
\begin{equation}
\bm{z}
=
\bm{W}\bm{x}
+
\bm{B}\bm{A}\bm{x},
\nonumber
\end{equation}
where $\bm{x}\in\mathbb{R}^{n}$ is the input activation,
$\bm{W}\in\mathbb{R}^{m\times n}$ is the frozen pretrained weight, and
$\bm{A}\in\mathbb{R}^{R\times n}$ and
$\bm{B}\in\mathbb{R}^{m\times R}$ are the input- and output-side factors,
respectively. An invertible change of coordinates in the latent space
can be absorbed into the two factors while preserving their product.
We therefore compare input subspaces through normalized representations.
The assumptions used in the analysis are stated below.

\begin{assumption}[Finite second moments]
\label{ass:finite-second-moments}
The activation $\bm{x}\in\mathbb{R}^{n}$ and the back-propagated
gradient $\bm{g}\in\mathbb{R}^{m}$ satisfy
\begin{equation}
    \mathbb{E}\!\left[
        \|\bm{x}\|_2^2
    \right]
    <
    \infty,
    \qquad
    \mathbb{E}\!\left[
        \|\bm{g}\|_2^2
        \|\bm{x}\|_2^2
    \right]
    <
    \infty.
    \nonumber
\end{equation}
Define the activation second-moment matrix as
\begin{equation}
    \bm{\Sigma}
    :=
    \mathbb{E}\!\left[
        \bm{x}\bm{x}^{\top}
    \right].
    \nonumber
\end{equation}
\end{assumption}

\begin{assumption}[Normalized subspace representation]
\label{ass:normalized-subspace}
The $R$-dimensional input subspace is represented by
$\bm{A}\in\mathbb{R}^{R\times n}$ with orthonormal rows:
\begin{equation}
    \bm{A}\bm{A}^{\top}
    =
    \bm{I}_{R},
    \qquad
    1
    \leq
    R
    <
    n.
    \nonumber
\end{equation}
\end{assumption}

Any full-row-rank factor admits a row-orthonormal representative with the
same row subspace. This normalization compares subspaces at a common
scale, and the activation-SVD basis used by \name has orthonormal rows
by construction.

\begin{assumption}[Standard zero-output initialization]
\label{ass:zero-output-initialization}
The loss $\ell(\bm{z},y)$ is differentiable with respect to $\bm{z}$,
and the output-side factor is initialized as $\bm{B}^{0}=\bm{0}$.
Define
\begin{equation}
    \bm{g}
    :=
    \left.
    \nabla_{\bm{z}}
    \ell(\bm{z},y)
    \right|_{\bm{z}=\bm{W}\bm{x}}.
    \nonumber
\end{equation}
\end{assumption}

\begin{assumption}[Covariance alignment]
\label{ass:covariance-alignment}
Define the gradient-weighted activation second moment as
\begin{equation}
    \bm{M}
    :=
    \mathbb{E}\!\left[
        \|\bm{g}\|_2^2
        \bm{x}\bm{x}^{\top}
    \right].
    \label{eq:gradient-weighted-second-moment}
\end{equation}
There exist $c>0$ and $\varepsilon\geq0$ such that
\begin{equation}
    \left\|
        \bm{M}
        -
        c\bm{\Sigma}
    \right\|_2
    \leq
    \varepsilon.
    \nonumber
\end{equation}
\end{assumption}

The quantity $\varepsilon$ bounds the change in activation geometry
induced by gradient weighting. This condition allows dependence between
$\bm{x}$ and $\bm{g}$ through the resulting second moments. For example,
$\mathbb{E}[\|\bm{g}\|_2^2\mid\bm{x}]=c$ implies
$\bm{M}=c\bm{\Sigma}$ even when $\bm{x}$ and $\bm{g}$ are dependent.

\subsubsection{Why Activation-SVD?}

For a normalized input subspace, define the captured activation energy
as
\begin{equation}
    \Phi(\bm{A})
    :=
    \mathbb{E}\!\left[
        \|\bm{A}\bm{x}\|_2^2
    \right].
    \nonumber
\end{equation}
Let
\begin{equation}
    \bm{\Sigma}
    =
    \bm{U}\operatorname{Diag}(\lambda_1,\ldots,\lambda_n)\bm{U}^{\top},
    \qquad
    \lambda_1\geq\cdots\geq\lambda_n\geq0,
    \nonumber
\end{equation}
and let $\bm{U}_{R}\in\mathbb{R}^{n\times R}$ contain the leading
$R$ eigenvectors. We define
$\bm{A}_{\mathrm{svd}}:=\bm{U}_{R}^{\top}$.
The classical principal-subspace variational principle underlies
activation-based initialization methods such as
EVA~\cite{paischer2026parametereva}. We state this activation-coverage
property together with its comparison to a uniformly random subspace.

\begin{theorem}[Optimal activation coverage]
\label{thm:optimal-activation-coverage}
Under Assumptions~\ref{ass:finite-second-moments} and
\ref{ass:normalized-subspace}, we have
\begin{equation}
    \Phi(\bm{A})
    =
    \operatorname{Tr}
    \left(
        \bm{A}
        \bm{\Sigma}
        \bm{A}^{\top}
    \right),
    \qquad
    \max_{\bm{A}\bm{A}^{\top}=\bm{I}_{R}}
    \Phi(\bm{A})
    =
    \sum_{i=1}^{R}\lambda_i.
    \nonumber
\end{equation}
The maximum is attained by $\bm{A}_{\mathrm{svd}}$. If
$\bm{A}_{\mathrm{rand}}$ represents a uniformly random
$R$-dimensional row subspace, then
\begin{equation}
    \mathbb{E}_{\bm{A}_{\mathrm{rand}}}
    \left[
        \Phi(\bm{A}_{\mathrm{rand}})
    \right]
    =
    \frac{R}{n}
    \operatorname{Tr}(\bm{\Sigma}).
    \nonumber
\end{equation}
Consequently,
\begin{equation}
    \Phi(\bm{A}_{\mathrm{svd}})
    \geq
    \mathbb{E}_{\bm{A}_{\mathrm{rand}}}
    \left[
        \Phi(\bm{A}_{\mathrm{rand}})
    \right].
    \nonumber
\end{equation}
The inequality is strict when $R<n$ and $\bm{\Sigma}$ is not a scalar
multiple of $\bm{I}_{n}$.
\end{theorem}

For \name, Theorem~\ref{thm:optimal-activation-coverage} motivates
allocating the input-subspace budget to the leading activation directions.
The random-subspace comparison describes the activation-energy benefit
of this choice when the input second moment is anisotropic.

At the initialization in
Assumption~\ref{ass:zero-output-initialization}, define the expected
gradient energy received by the output-side factor as
\begin{equation}
    \mathcal{G}(\bm{A})
    :=
    \mathbb{E}\!\left[
        \left\|
            \nabla_{\bm{B}}
            \ell
        \right\|_F^2
    \right].
    \nonumber
\end{equation}

\begin{theorem}[Initial gradient signal under covariance mismatch]
\label{thm:near-optimal-gradient-signal}
Under Assumptions~\ref{ass:finite-second-moments},
\ref{ass:normalized-subspace},
\ref{ass:zero-output-initialization}, and
\ref{ass:covariance-alignment}, we have
\begin{equation}
    \mathcal{G}(\bm{A})
    =
    \operatorname{Tr}
    \left(
        \bm{A}
        \bm{M}
        \bm{A}^{\top}
    \right).
    \nonumber
\end{equation}
Let $\bm{A}_{*}
\in
\operatorname*{arg\,max}_{
    \bm{A}\bm{A}^{\top}=\bm{I}_{R}
}
\mathcal{G}(\bm{A})$.
Then
\begin{equation}
    0
    \leq
    \mathcal{G}(\bm{A}_{*})
    -
    \mathcal{G}(\bm{A}_{\mathrm{svd}})
    \leq
    2R\varepsilon.
    \nonumber
\end{equation}
In particular, $\bm{A}_{\mathrm{svd}}$ is exactly optimal when
$\varepsilon=0$. If $\mathcal{G}(\bm{A}_{*})>0$, the relative gap satisfies
\begin{equation}
    \frac{
        \mathcal{G}(\bm{A}_{*})
        -
        \mathcal{G}(\bm{A}_{\mathrm{svd}})
    }{
        \mathcal{G}(\bm{A}_{*})
    }
    \leq
    \frac{2R\varepsilon}{\mathcal{G}(\bm{A}_{*})}.
    \nonumber
\end{equation}
Moreover,
\begin{equation}
    \mathbb{E}_{\bm{A}_{\mathrm{rand}}}
    \left[
        \mathcal{G}(\bm{A}_{\mathrm{rand}})
    \right]
    =
    \frac{R}{n}
    \operatorname{Tr}(\bm{M}).
    \nonumber
\end{equation}
\end{theorem}

Theorem~\ref{thm:near-optimal-gradient-signal} quantifies how covariance
mismatch controls the initial gradient-energy gap of activation-SVD.
Its relative form expresses this gap on the scale of the optimal initial
signal.

Appendix~\ref{app:lora_subspace_energy} provides complementary empirical
observations of activation coverage in learned LoRA input subspaces.
These results motivate a fixed input basis formed from the leading
activation directions. The next section combines this basis with a
grouped random output dictionary and learns layer-specific coefficients
within the resulting update subspace.

\section{\name}
\label{sec:method}

\name uses the activation geometry analyzed in
Section~\ref{sec:motivation} to construct a fixed input basis. Each input
direction is paired with a group of random output atoms, and adaptation
learns their combination coefficients. The groups define separate output
subspaces while keeping the effective update rank bounded by the number
of input directions. This construction keeps the trainable state compact
and supports efficient evaluation through a small set of effective
output vectors.

Consider a frozen linear layer
\(\bm{W}\in\mathbb{R}^{m\times n}\). \name
parameterizes its update as
\begin{equation}
    \Delta\bm{W}
    =
    s
    \sum_{r=1}^{R}
    \sum_{h=1}^{H}
    \gamma_{r,h}
    \bm{b}_{r,h}
    \bm{a}_{r}^{\top},
    \label{eq:scalar-update}
\end{equation}
where \(\bm{a}_{r}\in\mathbb{R}^{n}\) is a fixed task-aware input
direction, \(\bm{b}_{r,h}\in\mathbb{R}^{m}\) is a fixed random output
atom, \(\gamma_{r,h}\in\mathbb{R}\) is trainable, and \(s\) is a fixed
scaling factor. Each adapted layer contains \(K=RH\) trainable
coefficients. All coefficients are initialized at zero, so fine-tuning
starts from the pretrained model.

\subsection{Task-Aware Input Directions}

We collect \(S\) input activations from a small calibration set and stack
them into \(\bm{X}\in\mathbb{R}^{S\times n}\). Following the second-moment
analysis in Section~\ref{sec:motivation}, \(\bm{X}\) is not mean-centered.
Following activation-based initialization~\cite{paischer2026parametereva},
we compute the SVD and retain the leading $R$ right singular vectors:
\begin{equation}
    \bm{X}=\bm{U}\bm{D}\bm{V}^{\top},
    \qquad
    \bm{A}=\bm{V}_{:,1:R}^{\top}\in\mathbb{R}^{R\times n}.
    \nonumber
\end{equation}

The resulting directions form an orthonormal basis of the dominant
empirical activation subspace. We freeze this input basis during
fine-tuning, following fixed-input-factor designs
~\cite{zhang2023lorafa,lin2026tlora}. It defines the input coordinates
for coefficient learning over \name's fixed output dictionaries.

\subsection{Grouped and Shared Random Output Dictionaries}

For each input direction $\bm{a}_{r}$, we sample $H$ fixed output atoms and collect them into
\begin{equation}
    \bm{B}_{r}=\begin{bmatrix}\bm{b}_{r,1} & \cdots & \bm{b}_{r,H}\end{bmatrix}\in\mathbb{R}^{m\times H}.
    \nonumber
\end{equation}
The atoms are sampled once from a zero-mean isotropic distribution and
remain frozen. Let
$\bm{\gamma}_{r}=[\gamma_{r,1},\ldots,\gamma_{r,H}]^{\top}$ and
$\widetilde{\bm{b}}_{r}=\bm{B}_{r}\bm{\gamma}_{r}$. During training,
$\widetilde{\bm{b}}_{r}$ varies within the fixed span of $\bm{B}_{r}$.
Thus, each input direction has its own output subspace, and
\eqref{eq:scalar-update} becomes
\begin{equation}
    \Delta\bm{W}=s\sum_{r=1}^{R}\widetilde{\bm{b}}_{r}\bm{a}_{r}^{\top}=s\widetilde{\bm{B}}\bm{A}, \qquad \operatorname{rank}(\Delta\bm{W})\leq R.
    \nonumber
\end{equation}

Here, $R$ determines the input-subspace dimension and maximum update rank, while $H$ controls the number of candidate output directions associated with each input direction. Since the $H$ output atoms can be merged before multiplication with the activations, the token-dependent computation scales with $R$ rather than $K=RH$. Appendix~\ref{app:ablation_structure} provides the corresponding complexity analysis and empirical throughput comparison. For an input $\bm{x}$, the adapter output is
\begin{equation}
    \Delta\bm{W}\bm{x}=s\sum_{r=1}^{R}\left(\bm{a}_{r}^{\top}\bm{x}\right)\bm{B}_{r}\bm{\gamma}_{r}.
    \nonumber
\end{equation}
At zero coefficient initialization, isotropic random output atoms relate
the expected scalar-gradient energy to the gradient-weighted activation
energy $\mathcal{G}(\bm{A})$. The covariance-alignment condition connects
this quantity to activation coverage. Appendix~\ref{app:gradient_analysis}
provides the derivation for the trainable coefficients.

Each adapted layer maintains its own activation-derived basis $\bm{A}$
and trainable coefficient matrix $\bm{\Gamma}$. Layers of the same
module type with compatible output dimensions may share the fixed output
dictionaries $\{\bm{B}_{r}\}_{r=1}^{R}$. Layer-specific input bases and
coefficients produce distinct updates from the shared atoms. The sharing
ablation in Appendix~\ref{app:ablation_structure} shows lower memory
consumption with similar accuracy.

The construction reduces three sources of training cost. Learning only
the coefficient set reduces gradient and optimizer-state storage.
Grouped evaluation limits token-dependent adapter computation, while
dictionary sharing reduces the memory occupied by fixed output bases.
Section~\ref{sec:training_efficiency} reports the resulting peak memory
and throughput measurements. These properties are useful for on-device
personalization under limited memory and compute budgets.

\section{Experimental Results}

\begin{table*}[!t]
    \centering
    \caption{ \textbf{Commonsense reasoning accuracy of LLaMA3-8B and Qwen2.5-7B across 8 benchmarks.} Results are averaged over three independent runs. The \name configurations are highlighted. }
    \label{tab:accuracy_common_8b_7b}
    \resizebox{\linewidth}{!}{
    \begin{tabular}{l|l|c|cccccccc|c}
    \toprule
    \textbf{Model} & \textbf{Method} & \textbf{Params (\%)} & \textbf{BoolQ} & \textbf{PIQA} & \textbf{SIQA} & \textbf{HS} & \textbf{WG} & \textbf{ARC-E} & \textbf{ARC-C} & \textbf{OBQA} & \textbf{Avg.} \\
    \midrule
    \multirow{10}{*}{Qwen2.5-7B} & FT & 100.00 & 75.3 & 89.7 & 82.0 & 96.1 & 89.7 & 95.6 & 88.4 & 93.1 &  88.7$_{\pm 0.1}$\\
    \cmidrule{2-12}
    & LoRA ($r=1$) & 0.0331 & 74.3 & 90.3 & 81.9 & 95.8 & 88.7 & 96.1 & 89.2 & 91.7 & 88.5$_{\pm 0.1}$ \\
    & LoRA ($r=8$) & 0.2643 & 74.6 & 90.4 & 81.9 & 96.3 & 89.3 & 95.9 & 89.5 & 92.3 & 88.8$_{\pm 0.1}$ \\
    \cmidrule{2-12}
    & VeRA & 0.0208 & 73.7 & 90.2 & 81.5 & 95.6 & 87.7 & 96.3 & 88.8 & 91.5 & 88.2$_{\pm 0.2}$ \\
    & NOLA & 0.0105 & 73.6 & 90.0 & 82.0 & 95.8 & 88.6 & 96.4 & 89.2 & 91.5 & 88.4$_{\pm 0.0}$\\
    & LoRA-XS & 0.0105 & 73.4 & 89.8 & 82.4 & 95.6 & 87.2 & 96.5 & 88.5 & 90.9 & 88.0$_{\pm 0.2}$ \\
    & RandLoRA & 0.0175 & 73.9 & 89.5 & 81.2 & 95.8 & 88.3 & 96.1 & 89.1 & 90.8 & 88.1$_{\pm 0.2}$\\
    \cmidrule{2-12}
    \rowcolor{blue4}\cellcolor{white}& \name ($K=1024$) & 0.0026 & 73.4 & 90.3 & 82.4 & 96.0 & 87.5 & 96.8 & 89.0 & 92.3 & \textbf{88.5}$_{\pm 0.1}$ \\
    \rowcolor{blue3}\cellcolor{white}& \name ($K=2048$) & 0.0052 & 73.7 & 90.0 & 82.5 & 96.1 & 88.2 & 96.7 & 89.2 & 91.9 & \textbf{88.6}$_{\pm 0.0}$ \\
    \rowcolor{blue2}\cellcolor{white}& \name ($K=4096$) & 0.0105 & 74.0 & 90.3 & 82.2 & 96.2 & 89.3 & 96.5 & 89.5 & 93.0 & \textbf{88.9}$_{\pm 0.1}$\\
    \midrule
    \multirow{10}{*}{LLaMA3-8B} & FT & 100.00 & 76.2 & 90.4 & 83.0 & 96.7 & 89.4 & 94.1 & 85.1 & 88.8 & 88.0$_{\pm 0.2}$ \\
    \cmidrule{2-12}
    & LoRA ($r=1$) & 0.0326 & 74.3 & 89.9 & 81.9 & 96.3 & 89.5 & 93.3 & 83.3 & 89.0 & 87.2$_{\pm 0.2}$\\
    & LoRA ($r=8$) & 0.2604 & 75.9 & 90.0 & 83.0 & 96.6 & 89.4 & 93.3 & 83.1 & 87.8 & 87.4$_{\pm 0.3}$ \\
    \cmidrule{2-12}
    & VeRA & 0.0199 & 74.8 & 88.9 & 81.6 & 95.9 & 87.1 & 92.8 & 82.1 & 87.5 & 86.3$_{\pm 0.3}$\\
    & NOLA & 0.0114 & 74.7 & 89.5 & 82.8 & 96.4 & 88.3 & 92.9 & 82.9 & 87.4 & 86.9$_{\pm 0.1}$\\
    & LoRA-XS & 0.0114 & 73.3 & 88.3 & 82.7 & 96.0 & 88.1 & 92.4 & 81.9 & 87.7 & 86.2$_{\pm 0.2}$ \\
    & RandLoRA & 0.0203 & 75.1 & 89.5 & 82.3 & 95.5 & 86.7 & 93.2 & 81.7 & 88.4 & 86.6$_{\pm 0.0}$\\
    \cmidrule{2-12}
    \rowcolor{blue4}\cellcolor{white} & \name ($K=1024$) & 0.0028 & 73.5 & 89.2 & 82.4 & 96.2 & 88.0 & 92.8 & 82.3 & 87.6 & \textbf{86.5}$_{\pm 0.3}$\\
    \rowcolor{blue3}\cellcolor{white}& \name ($K=2048$) & 0.0057 & 74.2 & 89.7 & 82.5 & 96.3 & 87.5 & 93.3 & 83.5 & 88.3 & \textbf{86.9}$_{\pm 0.1}$\\
    \rowcolor{blue2}\cellcolor{white}& \name ($K=4096$) & 0.0114 & 74.0 & 89.6 & 82.8 & 96.8 & 88.8 & 93.1 & 83.9 & 89.2 & \textbf{87.3$_{\pm 0.1}$}\\
    \bottomrule
    \end{tabular}
    }
\end{table*} 

We evaluate adaptation quality at compact trainable budgets and measure
the corresponding training memory and throughput. The experiments cover
several language adaptation tasks. The Appendix includes additional diagnostic analyses, comparisons with enhanced LoRA variants, general instruction fine-tuning, larger-model evaluations, and extensive ablations on the parameterization, calibration, rank, initialization, and cross-layer sharing, together with full implementation details and parameter-budget comparisons.

\noindent\textbf{Training Configuration.}
We evaluate \name on commonsense reasoning with Qwen2.5~\cite{qwen25report}
and LLaMA3~\cite{grattafiori2024llama3}, and further test transfer to
mathematical reasoning and code generation. We also ablate the initialization
of the fixed bases $\bm{A}$ and $\bm{B}$.

\noindent\textbf{Baselines.}
We compare \name with full-parameter fine-tuning and five representative
PEFT methods: \textbf{Full Fine-tuning (FT)} as a full-capacity reference;
\textbf{LoRA}~\cite{hu2022lora}, the standard trainable low-rank
parameterization; \textbf{VeRA}~\cite{kopiczko2024vera}, which shares
random low-rank projections and learns layer-specific scaling vectors;
\textbf{LoRA-XS}~\cite{balazy2024loraxs}, which optimizes a compact matrix
between fixed projections; \textbf{NOLA}~\cite{abbasi2024nola}, which
learns combinations of fixed random bases; and
\textbf{RandLoRA}~\cite{albert2025randlora}, which aggregates random
low-rank bases with trainable diagonal scalings.

\subsection{Commonsense Reasoning Benchmarks}

We present the main accuracy results in Table~\ref{tab:accuracy_common_8b_7b} and the parameter--performance comparison in Figure~\ref{fig:common_param}. We fine-tune all methods on the Commonsense170K training set~\cite{hu2023llmadapters} and evaluate on 8 commonsense reasoning benchmarks. We conduct experiments on both LLaMA3-8B and Qwen2.5-7B, and report the mean accuracy over three independent runs. A more comprehensive comparison covering multiple trainable-parameter budgets for NOLA, LoRA-XS, and the other PEFT baselines is provided in Appendix Table~\ref{tab:accuracy_common_8b_7b_appendix}.

\noindent\textbf{Parameter Efficiency.}

Table~\ref{tab:accuracy_common_8b_7b} shows that \name approaches
LoRA-level accuracy with substantially smaller trainable budgets.
With $K=4096$, it reaches 88.9 average accuracy on Qwen2.5-7B and
87.3 on LLaMA3-8B, compared with 88.8 and 87.4 for rank-8 LoRA.
The corresponding trainable parameter counts are approximately $1/25$
and $1/23$ of LoRA's, respectively. Across the evaluated coefficient
budgets, \name records higher mean accuracy than LoRA-XS and remains
competitive with NOLA. Figure~\ref{fig:common_param} shows that the
compact coefficient parameterization sustains strong task performance
across the evaluated budgets.

\subsection{Training Memory and Throughput}
\label{sec:training_efficiency}

\begin{figure*}[!ht]
    \centering
    \begin{subfigure}[b]{0.48\linewidth}
        \centering
        \includegraphics[width=\linewidth]{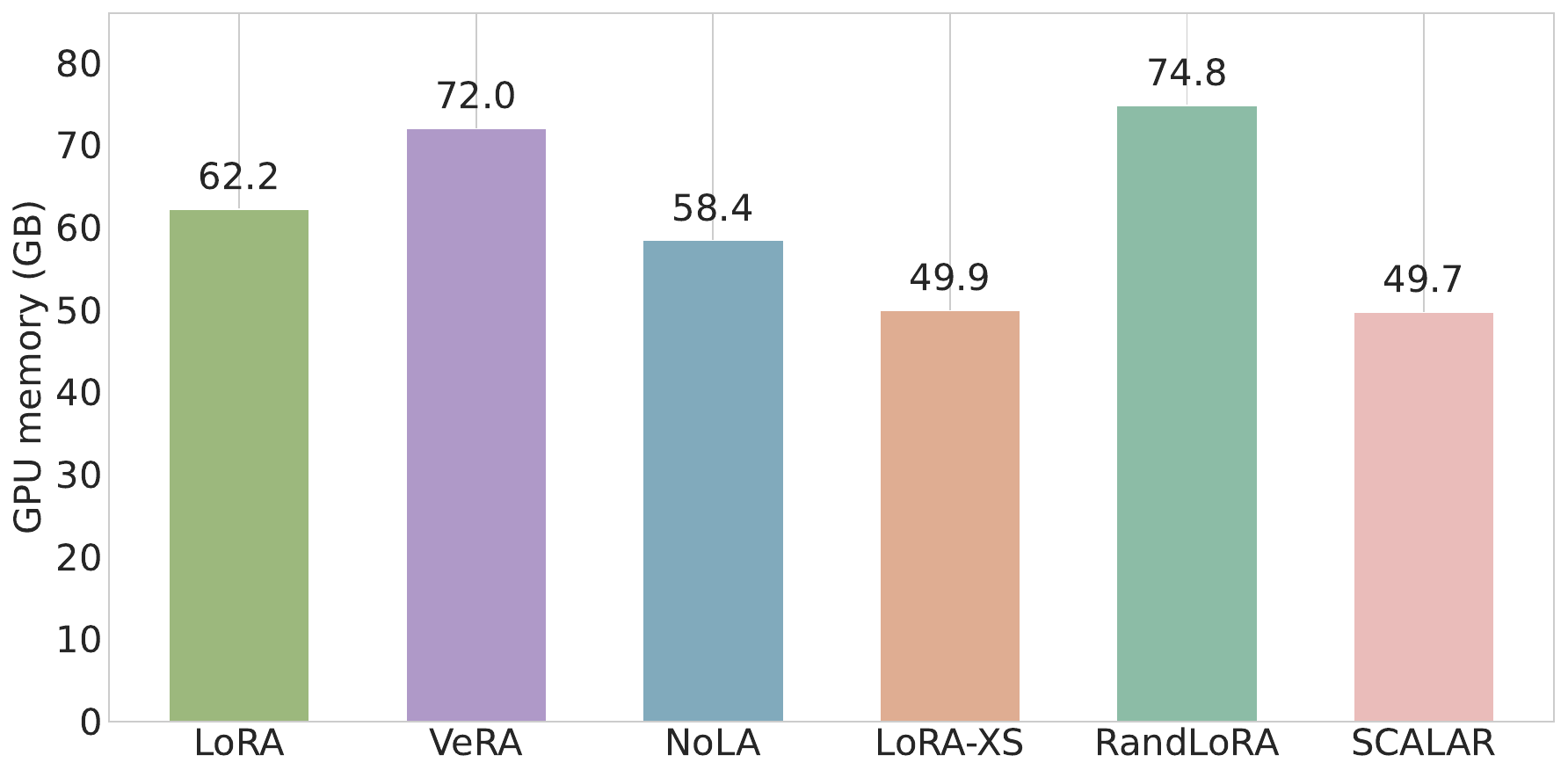}
        \caption{Memory comparison}
        \label{fig:mem_7b_common}
    \end{subfigure}
    \hfill
    \begin{subfigure}[b]{0.48\linewidth}
        \centering
        \includegraphics[width=\linewidth]{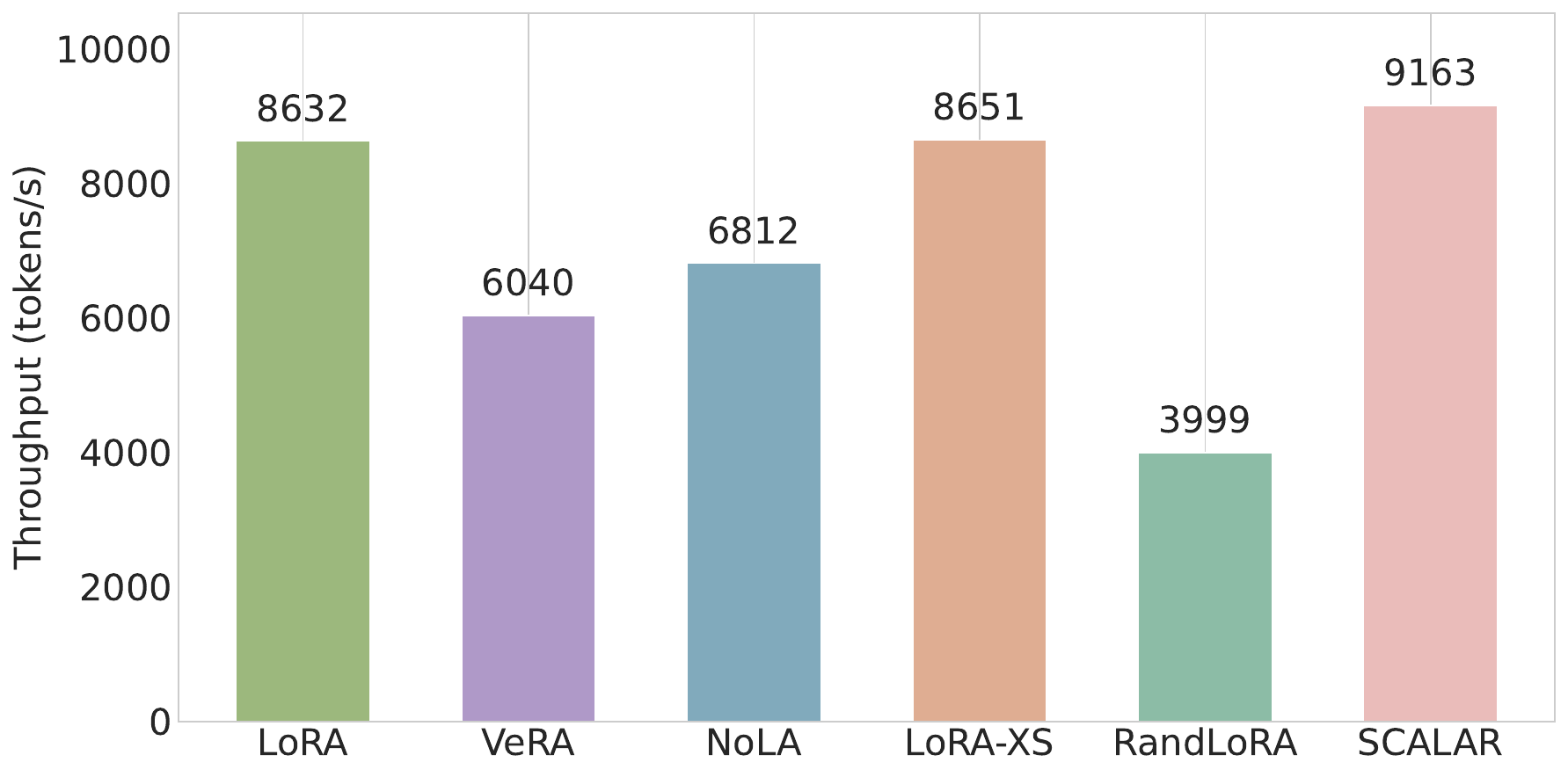}
        \caption{Training tokens/s}
        \label{fig:through_7b_common}
    \end{subfigure}
    \caption{\textbf{Memory and throughput for fine-tuning Qwen2.5-7B on commonsense reasoning.} We report peak GPU memory with a device batch size of 16 and token throughput on 2 NVIDIA H20 GPUs. \name uses less memory and achieves higher throughput than the compared baselines.}
    \label{fig:main}
\end{figure*}

Figure~\ref{fig:main} compares peak GPU memory and training token
throughput when fine-tuning Qwen2.5-7B. \name achieves the lowest
peak memory among the evaluated PEFT methods, using 20\% less memory
than LoRA, 15\% less than NOLA, and 33\% less than RandLoRA.
Training throughput increases by 6\% over LoRA, 34\% over NOLA, and
approximately 130\% over RandLoRA. Together with the accuracy results,
these measurements show that \name combines a compact trainable budget
with effective adaptation, lower peak memory, and higher throughput.
The structural ablations in Appendix~\ref{app:ablation_structure}
examine how grouped computation and cross-layer dictionary sharing
balance accuracy and training cost.

\subsection{Mathematical Reasoning and Code Generation}
We further evaluate the transferability of our method to mathematical reasoning and code generation tasks. For mathematical reasoning, all methods are trained on MetaMath~\cite{yu2023metamath} and evaluated on GSM8K~\cite{cobbe2021gsm8k} and MATH~\cite{hendrycks2021math}. For code generation, the models are trained on CodeFeedback~\cite{zheng2024codefeedback} and evaluated on HumanEval~\cite{chen2021humaneval} and MBPP~\cite{austin2021mbpp}. The results are summarized in Table~\ref{tab:math_qwen}.

Across both domains, \name remains competitive with full fine-tuning and
LoRA while using substantially fewer trainable parameters. The
$K=512$ configuration reaches a mathematical average of 66.9 with
0.0013\% trainable parameters, and $K=256$ reaches a code-generation
average of 79.4 with 0.0006\% trainable parameters. These results extend
the effectiveness of the fixed-subspace parameterization to
mathematical and code adaptation.

\begin{table}[!t]
    \centering
    \caption{
    \textbf{Comparison on mathematical and code generation benchmarks on Qwen2.5-7B.} \name achieves comparable performance on mathematical reasoning and code generation.
    }
    \resizebox{.9\linewidth}{!}{
    \begin{tabular}{l|c|ccc|ccc}
        \toprule
        \textbf{Method} & \textbf{Params (\%)} 
        & \textbf{GSM8K} & \textbf{MATH} & \textbf{Avg. }
        & \textbf{HumanEval} & \textbf{MBPP} & \textbf{Avg.} \\
        \midrule
        FT 
        & 100.00
        & 84.6 & {48.8} & 66.7 
        & 74.4 & 82.5 & 78.4 \\
        \midrule
        LoRA ($r=1$) 
        & 0.0331 
        & 84.8 & 48.3 & 66.6 
        & 74.4 & 81.5 & 78.0 \\
        LoRA ($r=8$) 
        & 0.2643 
        & 84.8 & 48.2 & 66.5 
        & 70.1 & 83.3 & 76.7 \\
        \midrule
        VeRA & 0.0189 & 84.4 & 48.7 & 66.6 & 72.6 & 82.5 & 77.6 \\
        NOLA & 0.0013 & 84.3 & 47.6 & 66.0 & 73.2 & 82.0 & 77.6 \\
        LoRA-XS & 0.0014 & 83.0 & 47.1 & 65.1 & 72.6 & 81.0 & 76.8 \\
        RandLoRA & 0.0175 & 84.3 & 48.0 & 66.2 & 72.0 & 84.4 & 78.2 \\
        \midrule
        \rowcolor{blue4}\name ($K=128$)
        & 0.0003 
        & 83.5 & 48.5 & 66.0 
        & 73.2 & 81.7 & 77.5 \\
        \rowcolor{blue3}\name ($K=256$)
        & 0.0006 
        & 84.3 & 48.5 & 66.4 
        & {76.8} & 82.0 & \textbf{79.4} \\
        \rowcolor{blue2}\name ($K=512$)
        & 0.0013 
        & {85.0} & 48.7 & \textbf{66.9} 
        & 73.8 & {84.1} & 79.0 \\
        \bottomrule
    \end{tabular}
    }
    \label{tab:math_qwen}
\end{table}

\subsection{On the Initialization Methods}
\label{app:ablation_initialization}

We investigate the effects of different initialization strategies for
the input-side basis $\bm{A}$ and the output-side dictionary $\bm{B}$.
As shown in Table~\ref{tab:initialization_ablation}, activation-SVD
consistently yields the strongest initialization for $\bm{A}$, outperforming
both random and weight-SVD initialization under either choice of $\bm{B}$.
This result supports our design of extracting $\bm{A}$ from the dominant
activation subspace, which better captures the downstream input geometry.

\begin{table}[!ht]
\centering
\caption{\textbf{Commonsense reasoning Acc. on Qwen2.5-3B with different initialization methods.}}
\label{tab:initialization_ablation}
\begin{tabular}{l|ccc}
\toprule
& Random-$\bm{A}$
& Weight-SVD-$\bm{A}$
& Activation-SVD-$\bm{A}$ \\
\midrule
Random-$\bm{B}$
& 84.3
& 84.4
& \textbf{85.2} \\
Weight-SVD-$\bm{B}$
& 84.5
& 84.6
& 85.1 \\
\bottomrule
\end{tabular}
\end{table}

With activation-SVD input directions, random and weight-SVD output
dictionaries achieve similar accuracy. Random output atoms additionally
support sharing across compatible layers, reducing the storage of
layer-specific output bases as examined in
Appendix~\ref{app:ablation_structure}. We therefore use activation-SVD
$\bm{A}$ and random, shareable $\bm{B}$ as the default construction.

\section{Conclusion}

We introduced \name, a PEFT method that combines competitive adaptation
performance with substantially fewer trainable parameters, lower peak
training memory, and higher throughput. The method learns layer-specific
scalar coefficients over
activation-derived input directions and grouped random output
dictionaries. Compact coefficient updates reduce the trainable state,
grouped evaluation limits token-dependent computation, and shared
dictionaries reduce fixed-basis storage. Our analysis relates the
task-calibrated input geometry to the initial learning signal, while
the experiments and structural ablations demonstrate its parameter and
training efficiency. These properties make \name
a promising approach for on-device personalization under limited
memory and compute budgets.

\bibliographystyle{plainnat}
\bibliography{reference}

\clearpage
\appendix
\onecolumn
\clearpage
\appendix

\startcontents[appendix]

\phantomsection
\section*{\centering Appendix Contents}
\addcontentsline{toc}{section}{Appendix Contents}

\vspace{0.5em}

\printcontents[appendix]{}{1}{%
\setcounter{tocdepth}{3}%
}

\clearpage

\section{Proofs and Supporting Analysis}
\label{app:motivation-proofs}

This section proves the activation-coverage and initial-gradient results in
Section~\ref{sec:motivation}. The supporting analysis then connects the
activation-derived input subspace to the gradients of the trainable
coefficients of \name.

\subsection{Proof of Theorem~\ref{thm:optimal-activation-coverage}}
\label{app:proof_activation_coverage}

\begin{proof}
Using $\mathbb{E}[\bm{x}\bm{x}^{\top}]=\bm{\Sigma}$ and cyclic invariance of the trace,
\begin{equation}
\begin{aligned}
    \Phi(\bm{A})
    &=
    \mathbb{E}\!\left[\|\bm{A}\bm{x}\|_2^2\right]
    =
    \mathbb{E}\!\left[
        \bm{x}^{\top}\bm{A}^{\top}\bm{A}\bm{x}
    \right] \\
    &=
    \mathbb{E}\!\left[
        \operatorname{Tr}\left(
            \bm{A}\bm{x}\bm{x}^{\top}\bm{A}^{\top}
        \right)
    \right]
    =
    \operatorname{Tr}\left(
        \bm{A}\bm{\Sigma}\bm{A}^{\top}
    \right).
\end{aligned}
\nonumber
\end{equation}

Let
\begin{equation}
    \bm{\Sigma}=\bm{U}\bm{\Lambda}\bm{U}^{\top},
    \qquad
    \bm{\Lambda}=\operatorname{Diag}(\lambda_1,\ldots,\lambda_n),
    \qquad
    \lambda_1\geq\cdots\geq\lambda_n\geq0,
    \nonumber
\end{equation}
and define $\bm{C}:=\bm{A}\bm{U}$. Since
$\bm{A}\bm{A}^{\top}=\bm{I}_R$ and $\bm{U}$ is orthogonal,
\begin{equation}
\begin{aligned}
    \bm{C}\bm{C}^{\top}
    &=
    \bm{A}\bm{U}\bm{U}^{\top}\bm{A}^{\top}
    =
    \bm{I}_R, \\
    \Phi(\bm{A})
    &=
    \operatorname{Tr}\left(
        \bm{C}\bm{\Lambda}\bm{C}^{\top}
    \right)
    =
    \sum_{i=1}^{n}\lambda_i\|\bm{c}_i\|_2^2,
\end{aligned}
\nonumber
\end{equation}
where $\bm{c}_i$ is the $i$-th column of $\bm{C}$.

Set
\begin{equation}
    q_i:=\|\bm{c}_i\|_2^2,
    \qquad
    \bm{P}:=\bm{C}^{\top}\bm{C}.
    \nonumber
\end{equation}
Then
\begin{equation}
\begin{aligned}
    \bm{P}^2
    &=
    \bm{C}^{\top}\bm{C}\bm{C}^{\top}\bm{C}
    =
    \bm{C}^{\top}\bm{C}
    =
    \bm{P}, \\
    q_i
    &=
    \bm{e}_i^{\top}\bm{P}\bm{e}_i,
    \qquad
    0\leq q_i\leq1, \\
    \sum_{i=1}^{n}q_i
    &=
    \operatorname{Tr}(\bm{P})
    =
    \operatorname{Tr}(\bm{C}\bm{C}^{\top})
    =
    R.
\end{aligned}
\nonumber
\end{equation}
Thus,
\begin{equation}
\begin{aligned}
    \Phi(\bm{A})
    &=
    \sum_{i=1}^{R}\lambda_iq_i
    +
    \sum_{i=R+1}^{n}\lambda_iq_i
    \leq
    \sum_{i=1}^{R}\lambda_iq_i
    +
    \lambda_R\sum_{i=R+1}^{n}q_i \\
    &=
    \sum_{i=1}^{R}\lambda_iq_i
    +
    \lambda_R\left(R-\sum_{i=1}^{R}q_i\right)
    =
    R\lambda_R+\sum_{i=1}^{R}(\lambda_i-\lambda_R)q_i \\
    &\leq
    R\lambda_R+\sum_{i=1}^{R}(\lambda_i-\lambda_R)
    =
    \sum_{i=1}^{R}\lambda_i.
\end{aligned}
\label{eq:activation-coverage-upper-bound}
\end{equation}

For $\bm{A}_{\mathrm{svd}}=\bm{U}_{R}^{\top}$,
\begin{equation}
\begin{aligned}
    \bm{A}_{\mathrm{svd}}\bm{\Sigma}\bm{A}_{\mathrm{svd}}^{\top}
    &=
    \operatorname{Diag}(\lambda_1,\ldots,\lambda_R), \\
    \Phi(\bm{A}_{\mathrm{svd}})
    &=
    \sum_{i=1}^{R}\lambda_i,
\end{aligned}
\nonumber
\end{equation}
so the bound in \eqref{eq:activation-coverage-upper-bound} is attained.

For the uniformly random $R$-dimensional row subspace, let
\begin{equation}
    \bm{P}_{\mathrm{rand}}
    :=
    \bm{A}_{\mathrm{rand}}^{\top}\bm{A}_{\mathrm{rand}}.
    \nonumber
\end{equation}
This is a rank-$R$ orthogonal projector. Rotational invariance gives, for every orthogonal $\bm{Q}\in\mathbb{R}^{n\times n}$,
\begin{equation}
\begin{aligned}
    \bm{P}_{\mathrm{rand}}
    &\overset{\mathrm{d}}{=}
    \bm{Q}^{\top}\bm{P}_{\mathrm{rand}}\bm{Q}, \\
    \bm{S}:=\mathbb{E}[\bm{P}_{\mathrm{rand}}]
    &=
    \bm{Q}^{\top}\bm{S}\bm{Q},
    \qquad
    \forall\,\bm{Q}\in O(n).
\end{aligned}
\nonumber
\end{equation}
Hence $\bm{S}=a\bm{I}_n$ for some scalar $a$, and
\begin{equation}
\begin{aligned}
    an
    =
    \operatorname{Tr}(\bm{S})
    =
    \mathbb{E}\left[
        \operatorname{Tr}(\bm{P}_{\mathrm{rand}})
    \right]
    =
    R,
    \text{and}\ 
    \mathbb{E}[\bm{P}_{\mathrm{rand}}]
    =
    \frac{R}{n}\bm{I}_n.
\end{aligned}
\nonumber
\end{equation}
Therefore,
\begin{equation}
\begin{aligned}
    \mathbb{E}_{\bm{A}_{\mathrm{rand}}}
    \left[\Phi(\bm{A}_{\mathrm{rand}})\right]
    &=
    \mathbb{E}\left[
        \operatorname{Tr}\left(
            \bm{A}_{\mathrm{rand}}
            \bm{\Sigma}
            \bm{A}_{\mathrm{rand}}^{\top}
        \right)
    \right] \\
    &=
    \mathbb{E}\left[
        \operatorname{Tr}\left(
            \bm{P}_{\mathrm{rand}}\bm{\Sigma}
        \right)
    \right] \\
    &=
    \operatorname{Tr}\left(
        \mathbb{E}[\bm{P}_{\mathrm{rand}}]\bm{\Sigma}
    \right) \\
    &=
    \frac{R}{n}\operatorname{Tr}(\bm{\Sigma})
    =
    \frac{R}{n}\sum_{i=1}^{n}\lambda_i.
\end{aligned}
\label{eq:random-activation-coverage}
\end{equation}

Finally, for $R<n$,
\begin{equation}
\begin{aligned}
    &\frac{1}{R}\sum_{i=1}^{R}\lambda_i
    \geq
    \frac{1}{n}\sum_{i=1}^{n}\lambda_i
    \geq
    \frac{1}{n-R}\sum_{i=R+1}^{n}\lambda_i
    \Longrightarrow
    \sum_{i=1}^{R}\lambda_i
    \geq
    \frac{R}{n}\sum_{i=1}^{n}\lambda_i.
\end{aligned}
\nonumber
\end{equation}
Together with \eqref{eq:activation-coverage-upper-bound} and
\eqref{eq:random-activation-coverage},
\begin{equation}
    \Phi(\bm{A}_{\mathrm{svd}})
    =
    \sum_{i=1}^{R}\lambda_i
    \geq
    \mathbb{E}_{\bm{A}_{\mathrm{rand}}}
    \left[\Phi(\bm{A}_{\mathrm{rand}})\right].
    \nonumber
\end{equation}

Equality for $R<n$ requires equality between the average of the first
$R$ eigenvalues and that of the remaining $n-R$ eigenvalues. Since
$\lambda_1\geq\cdots\geq\lambda_n$, this occurs only when
$\lambda_1=\cdots=\lambda_n$, equivalently
$\bm{\Sigma}=a\bm{I}_n$ for some $a\geq0$. Hence the comparison is
strict whenever $R<n$ and $\bm{\Sigma}$ is not a scalar multiple of
$\bm{I}_n$.

The proof is completed.
\end{proof}

\subsection{Proof of Theorem~\ref{thm:near-optimal-gradient-signal}}
\label{app:proof_gradient_signal}

\begin{proof}
At $\bm{B}=\bm{0}$, for an arbitrary perturbation $\mathrm{d}\bm{B}$,
\begin{equation}
\begin{aligned}
    \mathrm{d}\bm{z}
    &=
    \mathrm{d}\bm{B}\bm{A}\bm{x}, \\
    \mathrm{d}\ell
    &=
    \bm{g}^{\top}\mathrm{d}\bm{z}
    =
    \bm{g}^{\top}\mathrm{d}\bm{B}\bm{A}\bm{x}
    =
    \operatorname{Tr}\left(
        \bm{g}^{\top}\mathrm{d}\bm{B}\bm{A}\bm{x}
    \right)
    =
    \operatorname{Tr}\left[
        \left(\bm{g}(\bm{A}\bm{x})^{\top}\right)^{\top}
        \mathrm{d}\bm{B}
    \right].
\end{aligned}
\nonumber
\end{equation}
Hence
\begin{equation}
\begin{aligned}
    \nabla_{\bm{B}}\ell
    &=
    \bm{g}(\bm{A}\bm{x})^{\top}, \\
    \left\|\nabla_{\bm{B}}\ell\right\|_F^2
    &=
    \left\|\bm{g}(\bm{A}\bm{x})^{\top}\right\|_F^2
    =
    \|\bm{g}\|_2^2\|\bm{A}\bm{x}\|_2^2.
\end{aligned}
\nonumber
\end{equation}
Therefore, using
\eqref{eq:gradient-weighted-second-moment},
\begin{equation}
\begin{aligned}
    \mathcal{G}(\bm{A})
    &=
    \mathbb{E}\left[
        \left\|\nabla_{\bm{B}}\ell\right\|_F^2
    \right]
    =
    \mathbb{E}\left[
        \|\bm{g}\|_2^2\|\bm{A}\bm{x}\|_2^2
    \right]
    =
    \mathbb{E}\left[
        \|\bm{g}\|_2^2
        \bm{x}^{\top}\bm{A}^{\top}\bm{A}\bm{x}
    \right] \\
    &=
    \operatorname{Tr}\left(
        \bm{A}
        \mathbb{E}\left[
            \|\bm{g}\|_2^2\bm{x}\bm{x}^{\top}
        \right]
        \bm{A}^{\top}
    \right)
    =
    \operatorname{Tr}\left(
        \bm{A}\bm{M}\bm{A}^{\top}
    \right).
\end{aligned}
\nonumber
\end{equation}

Since the row-orthonormal feasible set is compact and
$\mathcal{G}$ is continuous, a maximizer $\bm{A}_{*}$ exists. Define
\begin{equation}
    \bm{E}:=\bm{M}-c\bm{\Sigma}.
    \nonumber
\end{equation}
By Assumption~\ref{ass:covariance-alignment},
$\|\bm{E}\|_2\leq\varepsilon$. Since $\bm{E}$ is symmetric, for any
feasible $\bm{A}$,
\begin{equation}
\begin{aligned}
    -\varepsilon\bm{I}_n
    &\preceq
    \bm{E}
    \preceq
    \varepsilon\bm{I}_n \\
    &\Longrightarrow
    -\varepsilon\bm{A}\bm{A}^{\top}
    \preceq
    \bm{A}\bm{E}\bm{A}^{\top}
    \preceq
    \varepsilon\bm{A}\bm{A}^{\top} \\
    &\Longrightarrow
    -\varepsilon\bm{I}_R
    \preceq
    \bm{A}\bm{E}\bm{A}^{\top}
    \preceq
    \varepsilon\bm{I}_R \\
    &\Longrightarrow
    \left|
        \operatorname{Tr}\left(
            \bm{A}\bm{E}\bm{A}^{\top}
        \right)
    \right|
    \leq
    R\varepsilon.
\end{aligned}
\label{eq:gradient-signal-perturbation-bound}
\end{equation}

Using $\bm{M}=c\bm{\Sigma}+\bm{E}$,
Theorem~\ref{thm:optimal-activation-coverage}, and
\eqref{eq:gradient-signal-perturbation-bound},
\begin{equation}
\begin{aligned}
    \mathcal{G}(\bm{A}_{*})
    -
    \mathcal{G}(\bm{A}_{\mathrm{svd}})
    &=
    c\left[
        \operatorname{Tr}\left(
            \bm{A}_{*}\bm{\Sigma}\bm{A}_{*}^{\top}
        \right)
        -
        \operatorname{Tr}\left(
            \bm{A}_{\mathrm{svd}}
            \bm{\Sigma}
            \bm{A}_{\mathrm{svd}}^{\top}
        \right)
    \right] \\
    &+
    \operatorname{Tr}\left(
        \bm{A}_{*}\bm{E}\bm{A}_{*}^{\top}
    \right)
    -
    \operatorname{Tr}\left(
        \bm{A}_{\mathrm{svd}}
        \bm{E}
        \bm{A}_{\mathrm{svd}}^{\top}
    \right) \\
    &\leq
    \operatorname{Tr}\left(
        \bm{A}_{*}\bm{E}\bm{A}_{*}^{\top}
    \right)
    -
    \operatorname{Tr}\left(
        \bm{A}_{\mathrm{svd}}
        \bm{E}
        \bm{A}_{\mathrm{svd}}^{\top}
    \right) \\
    &\leq
    2R\varepsilon.
\end{aligned}
\nonumber
\end{equation}
Since $\bm{A}_{*}$ maximizes $\mathcal{G}$,
\begin{equation}
    0
    \leq
    \mathcal{G}(\bm{A}_{*})
    -
    \mathcal{G}(\bm{A}_{\mathrm{svd}})
    \leq
    2R\varepsilon.
    \nonumber
\end{equation}
In particular, when $\varepsilon=0$,
\begin{equation}
    \mathcal{G}(\bm{A}_{*})
    =
    \mathcal{G}(\bm{A}_{\mathrm{svd}}),
    \nonumber
\end{equation}
so $\bm{A}_{\mathrm{svd}}$ is also a maximizer.

For a uniformly random $R$-dimensional row subspace, let
\begin{equation}
    \bm{P}_{\mathrm{rand}}
    :=
    \bm{A}_{\mathrm{rand}}^{\top}\bm{A}_{\mathrm{rand}}.
    \nonumber
\end{equation}
From the proof of Theorem~\ref{thm:optimal-activation-coverage},
\begin{equation}
    \mathbb{E}[\bm{P}_{\mathrm{rand}}]
    =
    \frac{R}{n}\bm{I}_n.
    \nonumber
\end{equation}
Thus,
\begin{equation}
\begin{aligned}
    \mathbb{E}_{\bm{A}_{\mathrm{rand}}}
    \left[\mathcal{G}(\bm{A}_{\mathrm{rand}})\right]
    &=
    \mathbb{E}\left[
        \operatorname{Tr}\left(
            \bm{A}_{\mathrm{rand}}
            \bm{M}
            \bm{A}_{\mathrm{rand}}^{\top}
        \right)
    \right]
    =
    \mathbb{E}\left[
        \operatorname{Tr}\left(
            \bm{P}_{\mathrm{rand}}\bm{M}
        \right)
    \right] \\
    &=
    \operatorname{Tr}\left(
        \mathbb{E}[\bm{P}_{\mathrm{rand}}]\bm{M}
    \right)
    =
    \frac{R}{n}\operatorname{Tr}(\bm{M}).
\end{aligned}
\nonumber
\end{equation}

When $\mathcal{G}(\bm{A}_{*})>0$, dividing the suboptimality bound by
$\mathcal{G}(\bm{A}_{*})$ gives
\begin{equation}
    \frac{
        \mathcal{G}(\bm{A}_{*})
        -
        \mathcal{G}(\bm{A}_{\mathrm{svd}})
    }{
        \mathcal{G}(\bm{A}_{*})
    }
    \leq
    \frac{2R\varepsilon}{\mathcal{G}(\bm{A}_{*})}.
    \nonumber
\end{equation}
This completes the proof.
\end{proof}

\subsection{Gradient Signal of the Scalar Coefficients}
\label{app:gradient_analysis}

We provide the detailed connection between the activation-derived input
basis and the optimization of the scalar coefficients.

For a single adapted linear layer, let
\(\bm{x}\in\mathbb{R}^{n}\) denote its input and
\(\bm{g}\in\mathbb{R}^{m}\) the gradient of the loss with respect to its
output. We analyze the local gradient contribution at zero-initialized
coefficients, \(\bm{\Gamma}=\bm{0}\). From
\eqref{eq:scalar-update}, the gradient of an individual coefficient is
\begin{equation}
    \frac{\partial\ell}{\partial\gamma_{r,h}}
    =
    s\left(\bm{b}_{r,h}^{\top}\bm{g}\right)
    \left(\bm{a}_{r}^{\top}\bm{x}\right).
    \nonumber
\end{equation}
Therefore, the gradient received by each scalar factorizes into the
activation along its input direction and the projection of the output
gradient onto its random output atom.

Let
\(\bm{\Gamma}\in\mathbb{R}^{R\times H}\) collect all trainable scalar
coefficients. Assume that the output atoms are independent of
\((\bm{x},\bm{g})\) and satisfy
\begin{equation}
    \mathbb{E}\left[
        \bm{b}_{r,h}\bm{b}_{r,h}^{\top}
    \right]
    =
    \sigma_b^2\bm{I}_{m}.
    \nonumber
\end{equation}
Define
\begin{equation}
    \mathcal{J}(\bm{A})
    :=
    \mathbb{E}_{\bm{B}}\mathbb{E}\left[
        \left\|\nabla_{\bm{\Gamma}}\ell\right\|_F^2
    \right].
    \nonumber
\end{equation}
The expected initial coefficient-gradient energy is
\begin{align}
    \mathcal{J}(\bm{A})
    &=
    s^2\sum_{r=1}^{R}\sum_{h=1}^{H}
    \mathbb{E}\left[
        \left(\bm{b}_{r,h}^{\top}\bm{g}\right)^2
        \left(\bm{a}_{r}^{\top}\bm{x}\right)^2
    \right]
    \nonumber\\
    &=
    s^2H\sigma_b^2
    \mathbb{E}\left[
        \|\bm{g}\|_2^2
        \sum_{r=1}^{R}
        \left(\bm{a}_{r}^{\top}\bm{x}\right)^2
    \right]
    \nonumber\\
    &=
    s^2H\sigma_b^2
    \mathbb{E}\left[
        \|\bm{g}\|_2^2\|\bm{A}\bm{x}\|_2^2
    \right]
    \nonumber\\
    &=
    s^2H\sigma_b^2\mathcal{G}(\bm{A}).
    \label{eq:scalar-gradient-energy}
\end{align}

\begin{figure}[!ht]
    \centering
    \includegraphics[width=0.5\linewidth]{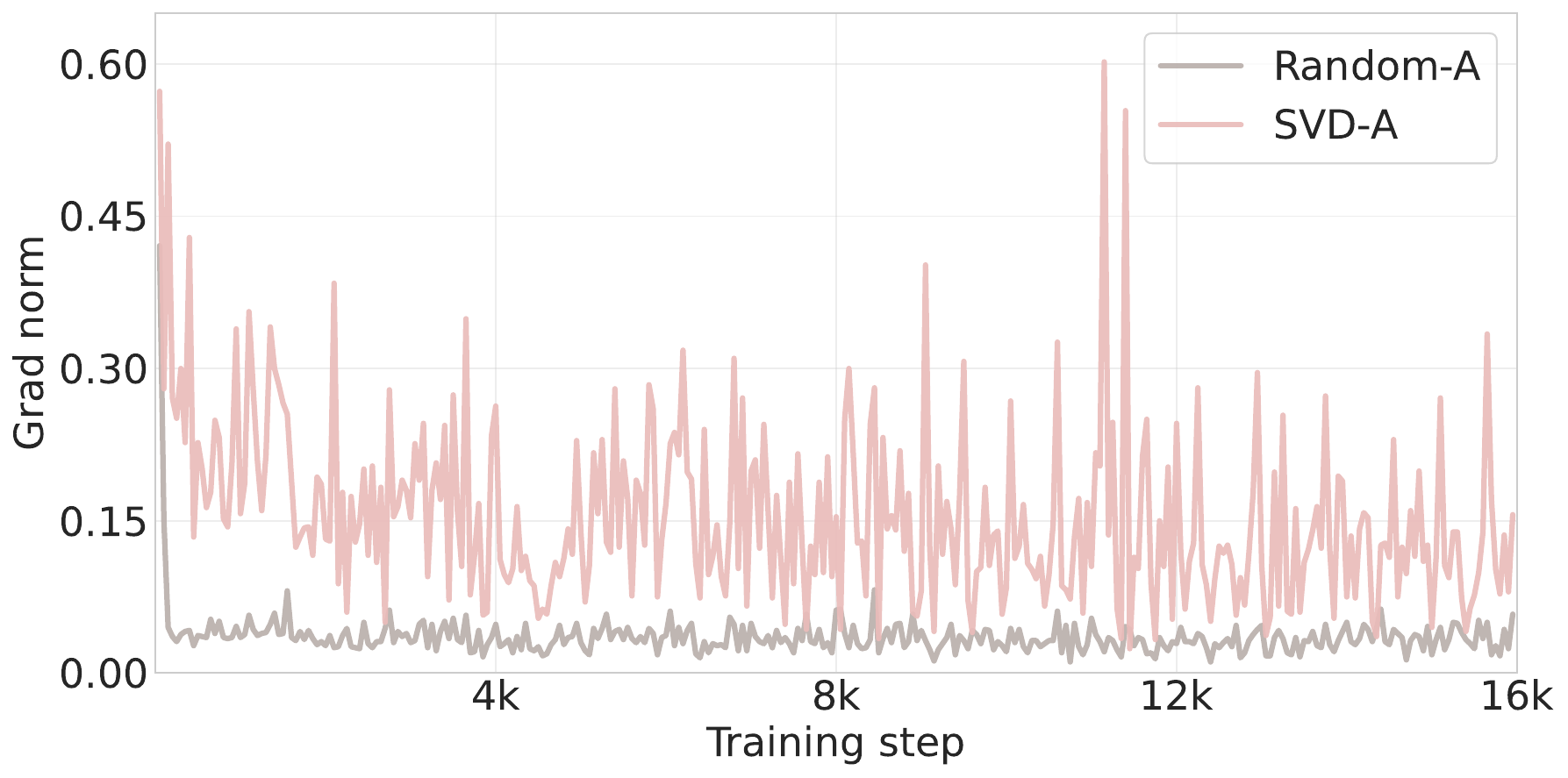}
    \caption{Gradient norms of the trainable adapter parameters when the
    input-side factor $\bm{A}$ uses activation-SVD or random initialization
    during Qwen2.5-7B fine-tuning on commonsense reasoning. The two input-side
    initializations are matched in scale. Activation-SVD initialization yields
    larger observed gradient norms over most of this run.}
    \label{fig:grad_norm}
\end{figure}

Under the stated isotropy and independence assumptions,
\eqref{eq:scalar-gradient-energy} gives the exact relation
\begin{equation}
    \mathcal{J}(\bm{A})
    =
    s^2H\sigma_b^2
    \operatorname{Tr}(\bm{A}\bm{M}\bm{A}^{\top}).
    \nonumber
\end{equation}
Thus, the expected initial coefficient-gradient energy is governed by the
gradient-weighted activation second moment \(\bm{M}\). When
\(\bm{M}=c\bm{\Sigma}\), it is proportional to the captured activation
energy \(\Phi(\bm{A})\). More generally, for the maximizer
\(\bm{A}_{*}\) in Theorem~\ref{thm:near-optimal-gradient-signal},
\begin{equation}
    0
    \leq
    \mathcal{J}(\bm{A}_{*})
    -
    \mathcal{J}(\bm{A}_{\mathrm{svd}})
    \leq
    2s^2H\sigma_b^2R\varepsilon.
    \nonumber
\end{equation}
This bound quantifies how the covariance-alignment error affects the
initial gradient energy of the coefficients of \name. The expectation measures
the squared norm of individual local gradient contributions.
Figure~\ref{fig:grad_norm} complements the analysis by comparing the
empirical gradients of the trainable adapter parameters under two
input-side initializations.

\section{Implementation Details}
\label{app:implementation_details}

This section supplements Section~\ref{sec:method} with implementation
details, training hyperparameters, baseline implementations, and the complete
commonsense reasoning results. Unless stated otherwise, all implementation
choices are shared across methods whenever their parameterizations permit.

\subsection{Implementation Optimizations}
\label{app:implementation}

We summarize the main implementation choices used in our experiments.
Notation follows Section~\ref{sec:method}. For each adapted layer,
\name maintains a task-aware input matrix
$\bm{A}\in\mathbb{R}^{R\times n}$ and a trainable coefficient matrix
$\bm{\Gamma}\in\mathbb{R}^{R\times H}$, where $R$ controls the maximum
rank of the aggregated update and $H$ is the number of random output
atoms per input direction. The random output dictionaries
$\{\bm{B}_{r}\}_{r=1}^{R}$ are frozen and may be shared across layers
with the same module type.

Coefficient-only optimization keeps adapter gradients and optimizer states
compact. Grouping reuses each input projection across its output atoms,
reducing token-dependent work. Cross-layer sharing limits the resident
storage of the frozen output dictionaries.

\subsubsection{Optional Sharing of Output Dictionaries}
\label{app:shared_output_dictionary}

Because the output atoms are sampled independently of pretrained
weights, compatible layers may reuse the same frozen dictionaries
$\{\bm{B}_{r}\}_{r=1}^{R}$. For example, query-projection layers with
the same output dimension can reuse one dictionary while retaining their
own activation-derived $\bm{A}$ and trainable $\bm{\Gamma}$. We use
this option in the default implementation to reduce the storage of frozen
random buffers. Appendix~\ref{app:ablation_structure} compares the shared
and unshared variants.

\subsubsection{Fast SVD for Input Calibration}
\label{app:fast_svd}

To construct $\bm{A}$, we collect input activations from a small
calibration set. Instead of storing the full activation matrix
$\bm{X}\in\mathbb{R}^{S\times n}$ on the accelerator, we accumulate
the Gram matrix
\begin{equation}
    \bm{C}=\bm{X}^{\top}\bm{X}
    \nonumber
\end{equation}
over streaming token blocks. We then extract the leading $R$ right
singular directions using randomized eigendecomposition
~\cite{halko2011finding}. This avoids dense SVD on $\bm{X}$ and is
efficient when $R\ll n$. Modules that receive the same hidden state
within a transformer layer, such as \texttt{q\_proj}, \texttt{k\_proj},
and \texttt{v\_proj}, share one calibration statistic and one
decomposition per layer.

To validate this Fast SVD against a dense (full) eigendecomposition of
$\bm{C}$, we compare both the wall-clock cost of the top-$R$ solve and
the recovered $R$-dimensional subspaces on the same Gram matrices
(Qwen2.5-7B; $R=16$; averaged over $84$ unique input groups).
Let $\bm{V}_{\mathrm{fast}},\bm{V}_{\mathrm{full}}\in\mathbb{R}^{n\times R}$
collect the top-$R$ right singular vectors, and let
$\bm{Q}_{\mathrm{fast}},\bm{Q}_{\mathrm{full}}$ be orthonormal bases of
their column spaces obtained by QR. The principal angles
$\{\theta_i\}_{i=1}^{R}$ satisfy
\begin{equation}
    \cos\theta_i
    =
    \sigma_i\!\left(
        \bm{Q}_{\mathrm{fast}}^{\top}\bm{Q}_{\mathrm{full}}
    \right),
    \qquad
    i=1,\ldots,R,
    \nonumber
\end{equation}
where $\sigma_i(\cdot)$ denotes the $i$-th singular value. We report the
mean cosine alignment
\begin{equation}
    \overline{\cos}
    =
    \frac{1}{R}\sum_{i=1}^{R}\cos\theta_i
    \label{eq:mean-cos-subspace}
\end{equation}
and the principal-angle affinity
$\frac{1}{R}\sum_{i=1}^{R}\cos^{2}\theta_i$.
In this implementation, Table~\ref{tab:fast-vs-full-svd} shows that the
randomized solver reduces the measured top-$R$ eigensolve time by
$19.6\times$ on the collected Gram matrices. The recovered subspaces are
also close under the reported principal-angle metrics
($\overline{\cos}=0.998$). The timing comparison covers the eigensolves
on these collected Gram matrices.

\begin{table}[!ht]
    \centering
    \caption{Fast (randomized) vs.\ full eigendecomposition for activation
    Gram $\bm{C}$ on Qwen2.5-7B ($R=16$).
    Time is the summed top-$R$ solve over all groups; subspace metrics compare
    Fast and full on the \emph{same} Grams.}
    \label{tab:fast-vs-full-svd}
    \begin{tabular}{lcc}
        \toprule
        & Fast SVD & Full SVD \\
        \midrule
        Eigendecomposition time (s) & $6.29$ & $123.0$ \\
        Speedup vs.\ full & $19.6\times$ & $1\times$ \\
        \midrule
        Mean cosine $\overline{\cos}$ \eqref{eq:mean-cos-subspace}
        & \multicolumn{2}{c}{$0.998$} \\
        Affinity $\frac{1}{R}\sum_i\cos^{2}\theta_i$
        & \multicolumn{2}{c}{$0.996$} \\
        Mean principal angle (deg)
        & \multicolumn{2}{c}{$0.500$} \\
        \bottomrule
    \end{tabular}
\end{table}

\subsubsection{Variance-Preserving Atom Scaling}
\label{app:atom_scaling}

\name represents each layer update as a sum of $K=RH$ rank-one atoms:
\begin{equation}
    \Delta\bm{W}
    =
    s\sum_{r=1}^{R}\sum_{h=1}^{H}
    \gamma_{r,h}\bm{b}_{r,h}\bm{a}_{r}^{\top}.
    \nonumber
\end{equation}
Here, the overall update has rank at most $R$, while each atom
$\bm{b}_{r,h}\bm{a}_{r}^{\top}$ has rank $1$. To keep the update scale
stable when $R$ or $H$ changes, we use
\begin{equation}
    s=\frac{\alpha}{\sqrt{K}},
    \qquad
    \alpha=1,
    \qquad
    K=RH.
    \label{eq:app_scale}
\end{equation}
We fix $\alpha=1$ across tasks and models. The factor $1/\sqrt{K}$
normalizes the aggregate scale of the $K$ output atoms.

The motivation is variance preservation. For fixed coefficients and input
projections of comparable scale, independent zero-mean output atoms with
matched second moments have vanishing cross terms in expectation. The
squared norm of the unscaled atom sum then grows linearly with $K$, up to
the coefficient- and activation-dependent factor:
\begin{equation}
    \mathbb{E}\left[
        \left\|
            \sum_{r=1}^{R}\sum_{h=1}^{H}
            \gamma_{r,h}\bm{b}_{r,h}
            \left(\bm{a}_{r}^{\top}\bm{x}\right)
        \right\|_2^2
    \right]
    =
    \mathcal{O}(K).
    \nonumber
\end{equation}
Under these conditions, the typical unscaled update norm grows as
$\mathcal{O}(\sqrt{K})$. Multiplication by $1/\sqrt{K}$ compensates for
this dependence on dictionary size.

The initial coefficient-gradient energy gives a second interpretation of
this scaling. At zero-initialized coefficients, the gradient of one scalar is
\begin{equation}
    \frac{\partial\ell}{\partial\gamma_{r,h}}
    =
    \frac{1}{\sqrt{K}}
    \left(\bm{b}_{r,h}^{\top}\bm{g}\right)
    \left(\bm{a}_{r}^{\top}\bm{x}\right).
    \nonumber
\end{equation}
Substituting $s=1/\sqrt{RH}$ into
\eqref{eq:scalar-gradient-energy} gives
\begin{equation}
    \mathcal{J}(\bm{A})
    =
    \frac{\sigma_b^2}{R}
    \operatorname{Tr}(\bm{A}\bm{M}\bm{A}^{\top}).
    \nonumber
\end{equation}
For fixed $R$ and output-atom variance, this expected initial gradient
energy is independent of $H$. Increasing $H$ therefore expands the output
dictionary while maintaining the same expected initial coefficient-gradient
energy under the stated assumptions.

\subsubsection{Efficient Adapter Forward Pass}
\label{app:efficient_forward}

The adapter output can be evaluated as
\begin{equation}
    \Delta\bm{W}\bm{x}
    =
    s\sum_{r=1}^{R}
    \left(\bm{a}_{r}^{\top}\bm{x}\right)
    \bm{B}_{r}\bm{\gamma}_{r}.
    \nonumber
\end{equation}
During training, we first compute
\begin{equation}
    \bm{h}=\bm{A}\bm{x},
    \nonumber
\end{equation}
then combine the random output atoms using the trainable coefficients
$\bm{\gamma}_{r}$. The resulting effective output vectors are reused
across the tokens in the batch. This keeps dictionary assembly outside the
token-dependent matrix products and avoids explicitly forming the full
update matrix $\Delta\bm{W}$ during training. Once training is finished, we
may cache the effective output directions
\begin{equation}
    \widetilde{\bm{b}}_{r}
    =
    \bm{B}_{r}\bm{\gamma}_{r}
    \nonumber
\end{equation}
or merge $\Delta\bm{W}$ into the pretrained weight, preserving the
inference-time behavior of standard LoRA-style adapters.

\subsection{Hyperparameters}
\label{app:hyperparameters}

We detail the hyperparameters used to reproduce our experimental
results. All methods are optimized with AdamW using
$\beta_1=0.9$, $\beta_2=0.999$, and
$\epsilon=1\mathrm{e}{-8}$. We use BF16 precision to reduce memory
consumption and train with Distributed Data Parallel (DDP). Gradient
synchronization is implemented using PyTorch's~\cite{paszke2017pytorch}
\textit{torch.distributed} framework. Following a widely used configuration for LLM adaptation, we apply all
PEFT methods to the same set of Transformer linear layers. This all-linear
setting provides sufficient adaptation capacity and ensures that differences
across methods arise from their parameterizations rather than from different
target-module selections.

Different PEFT methods have different preferred learning-rate ranges
and scaling conventions. For baselines with a LoRA-style scaling
factor, we use their default setting, namely
$\alpha=\operatorname{rank}$. For \name, we use the fixed
scaling in \eqref{eq:app_scale},
$s=1/\sqrt{RH}$, equivalently $\alpha/\sqrt{K}$ with
$\alpha=1$ and $K=RH$. This scaling is fixed across all tasks and
models.

For compactness, define the learning-rate search spaces as
\begin{equation}
    \mathcal{S}_{\mathrm{peft}}
    =
    \{
    1.0\mathrm{e}{-5},
    2.5\mathrm{e}{-5},
    5.0\mathrm{e}{-5},
    1.0\mathrm{e}{-4},
    2.5\mathrm{e}{-4},
    5.0\mathrm{e}{-4},
    1.0\mathrm{e}{-3}
    \},
    \nonumber
\end{equation}
\begin{equation}
    \mathcal{S}_{\mathrm{small}}
    =
    \{
    1.0\mathrm{e}{-6},
    2.5\mathrm{e}{-6},
    5.0\mathrm{e}{-6},
    1.0\mathrm{e}{-5},
    2.5\mathrm{e}{-5},
    5.0\mathrm{e}{-5},
    1.0\mathrm{e}{-4}
    \},
    \nonumber
\end{equation}
and
\begin{equation}
    \mathcal{S}_{\mathrm{large}}
    =
    \{
    1.0\mathrm{e}{-4},
    2.5\mathrm{e}{-4},
    5.0\mathrm{e}{-4},
    1.0\mathrm{e}{-3},
    2.5\mathrm{e}{-3},
    5.0\mathrm{e}{-3},
    1.0\mathrm{e}{-2}
    \}.
    \nonumber
\end{equation}
We tune LoRA, NOLA, LoRA-XS, and \name over
$\mathcal{S}_{\mathrm{peft}}$, RandLoRA and full-parameter
fine-tuning over $\mathcal{S}_{\mathrm{small}}$, and VeRA over
$\mathcal{S}_{\mathrm{large}}$.

\begin{table}[!ht]
    \centering
    \caption{Trainable params (\%) with different $K$ on Qwen2.5-7B and LLaMA3-8B.}
    \label{tab:trainable_params_k}
    \resizebox{0.8\linewidth}{!}{
    \begin{tabular}{l|ccc}
        \toprule
        Model & $K=128$ & $K=256$ & $K=512$ \\
        \midrule
        Qwen2.5-7B
        & 25,088 (0.0003\%)
        & 50,176 (0.0006\%)
        & 100,352 (0.0013\%) \\
        LLaMA3-8B
        & 28,672 (0.0003\%)
        & 57,344 (0.0006\%)
        & 114,688 (0.0014\%) \\
        \midrule
        & $K=1024$ & $K=2048$ & $K=4096$ \\
        \midrule
        Qwen2.5-7B
        & 200,704 (0.0026\%)
        & 401,408 (0.0052\%)
        & 802,816 (0.0105\%) \\
        LLaMA3-8B
        & 229,376 (0.0028\%)
        & 458,752 (0.0057\%)
        & 917,504 (0.0114\%) \\
        \bottomrule
    \end{tabular}
    }
\end{table}

All methods are fine-tuned for three epochs with batch size $32$ and a
cosine learning-rate scheduler. The first $100$ steps are used for
warm-up. We report results over three random seeds, $(0,1,2)$. The
maximum sequence length is set to $384$, which is sufficient for
commonsense reasoning inputs. For our method, we evaluate three default settings: $K$ in $\{1024,2048,4096\}$. A larger $K$ corresponds to higher model capacity and a larger number of trainable parameters. The trainable parameter counts for different values of $K$ are reported in Table~\ref{tab:trainable_params_k}.
Table~\ref{tab:hyper_commonsense}
summarizes the main hyperparameter settings.

\begin{table*}[!t]
    \centering
    \caption{Hyperparameters for fine-tuning LLaMA3-8B and Qwen2.5-7B on commonsense reasoning tasks.}
    \label{tab:hyper_commonsense}
    \setlength{\tabcolsep}{5pt}
    \renewcommand{\arraystretch}{1.15}
    \resizebox{\linewidth}{!}{
    \begin{tabular}{l|ccccccc}
        \toprule
        Hyperparameters
        & FT & LoRA & VeRA & NOLA & LoRA-XS & RandLoRA & \name \\
        \midrule
        Optimizer & \multicolumn{7}{c}{AdamW} \\
        LR scheduler & \multicolumn{7}{c}{Cosine} \\
        Precision & \multicolumn{7}{c}{BF16} \\
        Batch size & \multicolumn{7}{c}{$32$} \\
        Epochs & \multicolumn{7}{c}{$3$} \\
        Warmup steps & \multicolumn{7}{c}{$100$} \\
        Seeds & \multicolumn{7}{c}{$\{0,1,2\}$} \\
        Max sequence length & \multicolumn{7}{c}{$384$} \\
        Target modules & \multicolumn{7}{c}{All linear projection modules} \\
        \midrule
        Rank $(r)$
        & -
        & $\{1,8\}$
        & $\{1024,2048\}$
        & $\{1,8\}$
        & $\{32,48,64\}$
        & $\{2048,4096\}$
        & $\{8,16\}$ \\
        Scaling $(\alpha)$
        & - & $r$ & Default & $r$ & $r$ & $r$ & \eqref{eq:app_scale} \\
        LR (Qwen)
        & $2.5\mathrm{e}{-5}$
        & $1.0\mathrm{e}{-4}$
        & $2.5\mathrm{e}{-3}$
        & $1.0\mathrm{e}{-3}$
        & $5.0\mathrm{e}{-4}$
        & $2.5\mathrm{e}{-5}$
        & $5.0\mathrm{e}{-4}$ \\
        LR (LLaMA)
        & $1.0\mathrm{e}{-5}$
        & $1.0\mathrm{e}{-4}$
        & $2.5\mathrm{e}{-3}$
        & $5.0\mathrm{e}{-4}$
        & $5.0\mathrm{e}{-4}$
        & $2.5\mathrm{e}{-5}$
        & $5.0\mathrm{e}{-4}$ \\
        \bottomrule
    \end{tabular}
    }
\end{table*}

For mathematical reasoning and code generation, we fine-tune on
MetaMath~\cite{yu2023metamath} and CodeFeedback
~\cite{zheng2024codefeedback}, respectively, for one epoch. We use a batch
size of $256$, a maximum sequence length of $1024$, a $100$-step warm-up,
and cosine learning-rate decay.
The complete hyperparameter settings are provided in Table~\ref{tab:hyper_math_code}.

\begin{table}[!ht]
    \centering
    \caption{Hyperparameters for fine-tuning Qwen2.5-7B on mathematical reasoning and code generation tasks.}
    \label{tab:hyper_math_code}
    \setlength{\tabcolsep}{5pt}
    \renewcommand{\arraystretch}{1.15}
    \resizebox{\linewidth}{!}{
    \begin{tabular}{l|ccccccc}
        \toprule
        Hyperparameters
        & FT & LoRA & VeRA & NOLA & LoRA-XS & RandLoRA & \name \\
        \midrule
        LR scheduler & \multicolumn{7}{c}{Cosine} \\
        Precision & \multicolumn{7}{c}{BF16} \\
        Batch size & \multicolumn{7}{c}{$256$} \\
        Epochs & \multicolumn{7}{c}{$1$} \\
        Warmup steps & \multicolumn{7}{c}{$100$} \\
        Max sequence length & \multicolumn{7}{c}{$1024$} \\
        Target modules & \multicolumn{7}{c}{All linear projection modules} \\
        \midrule
        Rank $(r)$
        & - & $\{1,8\}$ & $256$ & $1$ & $24$ & $4096$ & $16$ \\
        Scaling $(\alpha)$
        & - & $r$ & Default & $r$ & $r$ & $r$ & \eqref{eq:app_scale} \\
        LR (Math)
        & $5.0\mathrm{e}{-6}$
        & $1.0\mathrm{e}{-5}$
        & $2.5\mathrm{e}{-4}$
        & $5.0\mathrm{e}{-5}$
        & $2.5\mathrm{e}{-4}$
        & $2.5\mathrm{e}{-5}$
        & $2.5\mathrm{e}{-5}$ \\
        LR (Code)
        & $2.5\mathrm{e}{-6}$
        & $\{1.0\mathrm{e}{-5},1.0\mathrm{e}{-4}\}$
        & $5.0\mathrm{e}{-4}$
        & $1.0\mathrm{e}{-4}$
        & $1.0\mathrm{e}{-4}$
        & $2.5\mathrm{e}{-5}$
        & $5.0\mathrm{e}{-4}$ \\
        \bottomrule
    \end{tabular}
    }
\end{table}

\subsection{Baseline Implementations}
\label{app:baseline_implementations}

All experiments use a common training stack based on Hugging Face
\texttt{transformers} and \texttt{peft} (v0.18.0), launched with
\texttt{torchrun} and the Hugging Face \texttt{Trainer}. LoRA
~\cite{hu2022lora} and VeRA~\cite{kopiczko2024vera} use the standard PEFT
APIs (\texttt{LoraConfig} and \texttt{VeraConfig}). RandLoRA
~\cite{albert2025randlora} uses the upstream \texttt{RandLoraConfig}
integration. LoRA-XS~\cite{balazy2024loraxs} is implemented on top of PEFT
LoRA with SVD-derived frozen $\bm{A}$ and $\bm{B}$ and a trainable
$r\times r$ matrix $\bm{M}$. NOLA~\cite{abbasi2024nola} uses an
in-house adapter module checked against the reference parameterization
because it is not available as a native PEFT tuner. Commonsense reasoning
uses generative decoding with regex-based answer extraction following the
DoRA/LoRA-XS evaluation style; mathematical reasoning follows the PiSSA
setup~\cite{meng2024pissa}; and code generation uses the corresponding
prompts with \texttt{evalplus} pass@1 on the base test suites.

\subsection{Complete Commonsense Reasoning Results}
\label{app:supp_commonsense}

The main text reports a compact comparison at representative parameter
budgets. Table~\ref{tab:accuracy_common_8b_7b_appendix} provides the
corresponding per-task results and variation across runs for the evaluated
budgets.

At the largest reported coefficient budgets, \name obtains average
accuracies of 88.9 on Qwen2.5-7B and 87.3 on LLaMA3-8B, compared with
88.8 and 87.4 for rank-8 LoRA. The trainable-parameter percentages are
0.0105\% and 0.0114\% for \name, compared with 0.2643\% and 0.2604\%
for LoRA. These results show comparable accuracy on both backbones with
substantially smaller trainable-parameter budgets.

\begin{table*}[!t]
    \centering
    \caption{\textbf{Accuracy comparison (3 independent runs) of Qwen2.5-7B and LLaMA3-8B with various PEFT methods on eight commonsense reasoning benchmarks.} This table supplements the main commonsense reasoning table by reporting the performance of each baseline under different trainable parameter budgets.}
    \label{tab:accuracy_common_8b_7b_appendix}
    \resizebox{\linewidth}{!}{
    \begin{tabular}{l|l|c|cccccccc|c}
        \toprule
        \textbf{Model} & \textbf{Method} & \textbf{Params (\%)} & \textbf{BoolQ} & \textbf{PIQA} & \textbf{SIQA} & \textbf{HS} & \textbf{WG} & \textbf{ARC-E} & \textbf{ARC-C} & \textbf{OBQA} & \textbf{Avg.} \\
        \midrule
        \multirow{15}{*}{Qwen2.5-7B} & FT & 100.00 & 75.3 & 89.7 & 82.0 & 96.1 & 89.7 & 95.6 & 88.4 & 93.1 & 88.7$_{\pm 0.1}$ \\
        \cmidrule{2-12}
        & LoRA ($r=1$) & 0.0331 & 74.3 & 90.3 & 81.9 & 95.8 & 88.7 & 96.1 & 89.2 & 91.7 & 88.5$_{\pm 0.1}$ \\
        & LoRA ($r=8$) & 0.2643 & 74.6 & 90.4 & 81.9 & 96.3 & 89.3 & 95.9 & 89.5 & 92.3 & 88.8$_{\pm 0.1}$ \\
        \cmidrule{2-12}
        & VeRA & 0.0208 & 73.7 & 90.2 & 81.5 & 95.6 & 87.7 & 96.3 & 88.8 & 91.5 & 88.2$_{\pm 0.2}$ \\
        & VeRA & 0.0235 & 73.6 & 90.0 & 81.8 & 95.8 & 88.0 & 96.1 & 89.5 & 91.2 & 88.2$_{\pm 0.1}$ \\
        \cmidrule{2-12}
        & NOLA & 0.0026 & 73.2 & 89.3 & 82.5 & 95.8 & 88.0 & 96.5 & 89.0 & 91.2 & 88.2$_{\pm 0.1}$ \\
        & NOLA & 0.0052 & 73.3 & 89.6 & 82.1 & 95.8 & 88.8 & 96.6 & 89.1 & 91.3 & 88.3$_{\pm 0.2}$ \\
        & NOLA & 0.0105 & 73.6 & 90.0 & 82.0 & 95.8 & 88.6 & 96.4 & 89.2 & 91.5 & 88.4$_{\pm 0.0}$ \\
        \cmidrule{2-12}
        & LoRA-XS & 0.0026 & 72.5 & 89.2 & 82.5 & 95.4 & 86.1 & 96.1 & 89.4 & 91.5 & 87.8$_{\pm 0.2}$ \\
        & LoRA-XS & 0.0052 & 72.6 & 89.6 & 82.7 & 95.5 & 86.5 & 96.5 & 88.5 & 91.2 & 87.9$_{\pm 0.1}$ \\
        & LoRA-XS & 0.0105 & 73.4 & 89.8 & 82.4 & 95.6 & 87.2 & 96.5 & 88.5 & 90.9 & 88.0$_{\pm 0.2}$ \\
        \cmidrule{2-12}
        & RandLoRA & 0.0175 & 73.9 & 89.5 & 81.2 & 95.8 & 88.3 & 96.1 & 89.1 & 90.8 & 88.1$_{\pm 0.2}$ \\
        & RandLoRA & 0.0225 & 73.6 & 89.6 & 81.0 & 95.7 & 88.3 & 96.2 & 89.2 & 91.8 & 88.2$_{\pm 0.2}$ \\
        \cmidrule{2-12}
        \rowcolor{blue4}\cellcolor{white} & \name ($K=1024$) & 0.0026 & 73.4 & 90.3 & 82.4 & 96.0 & 87.5 & 96.8 & 89.0 & 92.3 & \textbf{88.5}$_{\pm 0.1}$ \\
        \rowcolor{blue3}\cellcolor{white} & \name ($K=2048$) & 0.0052 & 73.7 & 90.0 & 82.5 & 96.1 & 88.2 & 96.7 & 89.2 & 91.9 & \textbf{88.6}$_{\pm 0.0}$ \\
        \rowcolor{blue2}\cellcolor{white} & \name ($K=4096$) & 0.0105 & 74.0 & 90.3 & 82.2 & 96.2 & 89.3 & 96.5 & 89.5 & 93.0 & \textbf{88.9}$_{\pm 0.1}$ \\
        \midrule
        \multirow{15}{*}{LLaMA3-8B} & FT & 100.00 & 76.2 & 90.4 & 83.0 & 96.7 & 89.4 & 94.1 & 85.1 & 88.8 & 88.0$_{\pm 0.2}$ \\
        \cmidrule{2-12}
        & LoRA ($r=1$) & 0.0326 & 74.3 & 89.9 & 81.9 & 96.3 & 89.5 & 93.3 & 83.3 & 89.0 & 87.2$_{\pm 0.2}$ \\
        & LoRA ($r=8$) & 0.2604 & 75.9 & 90.0 & 83.0 & 96.6 & 89.4 & 93.3 & 83.1 & 87.8 & 87.4$_{\pm 0.3}$ \\
        \cmidrule{2-12}
        & VeRA & 0.0199 & 74.8 & 88.9 & 81.6 & 95.9 & 87.1 & 92.8 & 82.1 & 87.5 & 86.3$_{\pm 0.3}$ \\
        & VeRA & 0.0228 & 75.3 & 89.0 & 82.4 & 95.7 & 88.1 & 93.3 & 82.1 & 85.5 & 86.4$_{\pm 0.0}$ \\
        \cmidrule{2-12}
        & NOLA & 0.0028 & 74.0 & 89.1 & 81.8 & 96.2 & 87.8 & 93.0 & 82.7 & 88.3 & 86.6$_{\pm 0.3}$ \\
        & NOLA & 0.0057 & 74.7 & 89.5 & 82.5 & 96.4 & 88.4 & 93.0 & 83.2 & 88.2 & 87.0$_{\pm 0.3}$ \\
        & NOLA & 0.0114 & 74.7 & 89.5 & 82.8 & 96.4 & 88.3 & 92.9 & 82.9 & 87.4 & 86.9$_{\pm 0.1}$ \\
        \cmidrule{2-12}
        & LoRA-XS & 0.0028 & 73.0 & 88.7 & 81.2 & 95.5 & 86.0 & 91.9 & 80.7 & 84.5 & 85.1$_{\pm 0.0}$ \\
        & LoRA-XS & 0.0064 & 73.0 & 88.6 & 82.5 & 95.9 & 86.9 & 92.5 & 81.7 & 87.1 & 86.0$_{\pm 0.0}$ \\
        & LoRA-XS & 0.0114 & 73.3 & 88.3 & 82.7 & 96.0 & 88.1 & 92.4 & 81.9 & 87.7 & 86.2$_{\pm 0.2}$ \\
        \cmidrule{2-12}
        & RandLoRA & 0.0203 & 75.1 & 89.5 & 82.3 & 95.5 & 86.7 & 93.2 & 81.7 & 88.4 & 86.6$_{\pm 0.0}$ \\
        & RandLoRA & 0.0269 & 75.4 & 89.1 & 82.3 & 95.6 & 87.2 & 92.9 & 82.5 & 87.1 & 86.5$_{\pm 0.2}$ \\
        \cmidrule{2-12}
        \rowcolor{blue4}\cellcolor{white} & \name ($K=1024$) & 0.0028 & 73.5 & 89.2 & 82.4 & 96.2 & 88.0 & 92.8 & 82.3 & 87.6 & \textbf{86.5}$_{\pm 0.3}$ \\
        \rowcolor{blue3}\cellcolor{white} & \name ($K=2048$) & 0.0057 & 74.2 & 89.7 & 82.5 & 96.3 & 87.5 & 93.3 & 83.5 & 88.3 & \textbf{86.9}$_{\pm 0.1}$ \\
        \rowcolor{blue2}\cellcolor{white} & \name ($K=4096$) & 0.0114 & 74.0 & 89.6 & 82.8 & 96.8 & 88.8 & 93.1 & 83.9 & 89.2 & \textbf{87.3$_{\pm 0.1}$} \\
        \bottomrule
    \end{tabular}
    }
\end{table*}

\section{Additional Results and Ablations}
\label{app:additional_results}

This section provides the diagnostic experiments and ablations referenced in
the main text. The results examine direction reweighting, activation
coverage, and the accuracy and computational cost of the proposed
parameterization.

\subsection{Learned-Direction Reweighting}
\label{app:direction_reweighting}
\label{app:oracle_directions}

This subsection expands the learned-direction reweighting diagnostic introduced
in Section~\ref{sec:motivation}. For each adapted matrix, we first train a
rank-one LoRA adapter, freeze its learned input and output directions, discard
the learned magnitude, and initialize a new scalar coefficient. The second
stage optimizes only these scalars using the same data and training protocol as
the rank-one LoRA run. This construction examines coefficient reweighting
with task-informed directions obtained from the trained adapter.

\begin{figure}[!t]
    \centering
    \begin{subfigure}[b]{0.45\linewidth}
        \centering
        \includegraphics[width=\linewidth]{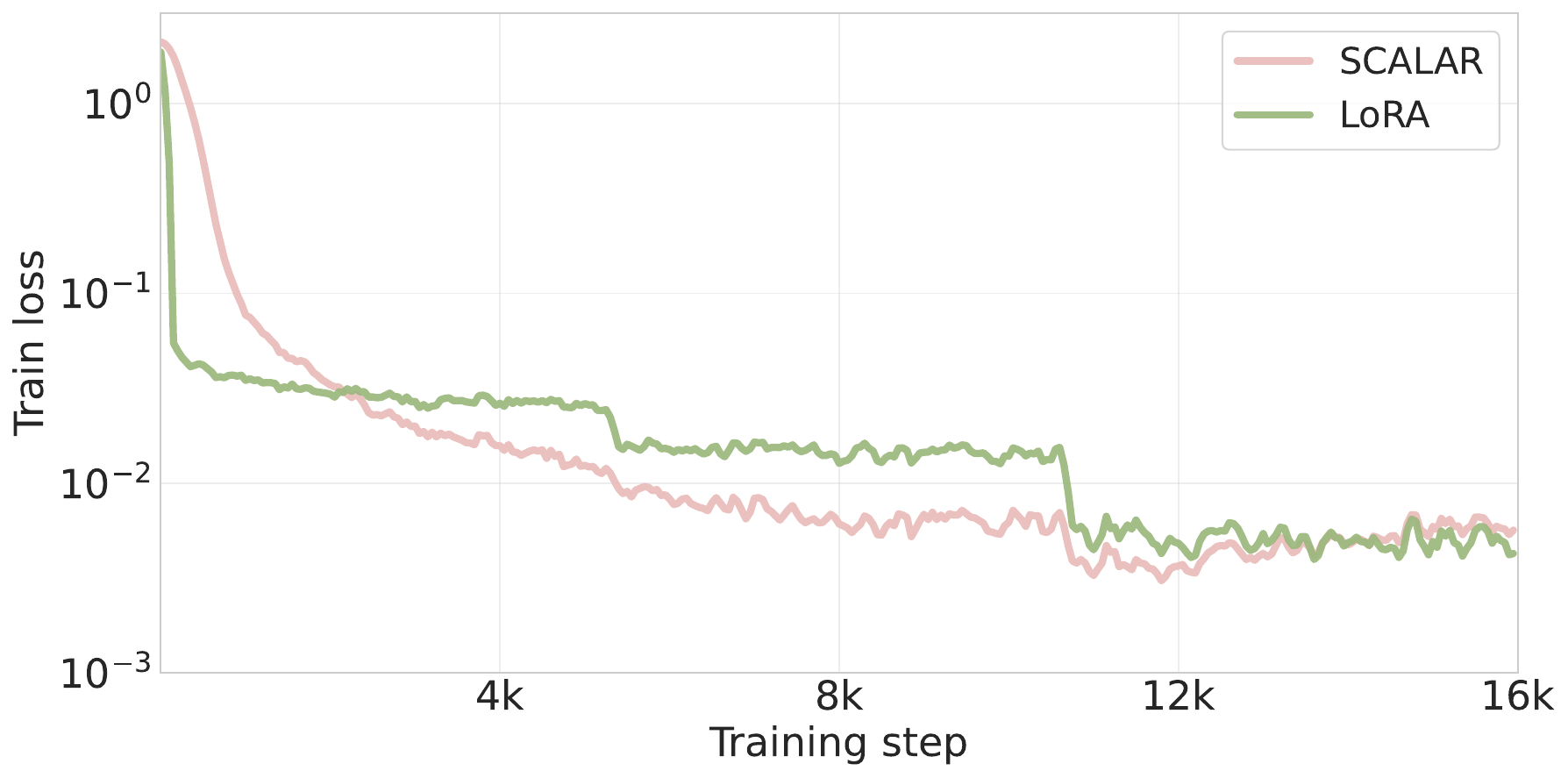}
        \caption{Qwen2.5-7B}
    \end{subfigure}
    \hfill
    \begin{subfigure}[b]{0.45\linewidth}
        \centering
        \includegraphics[width=\linewidth]{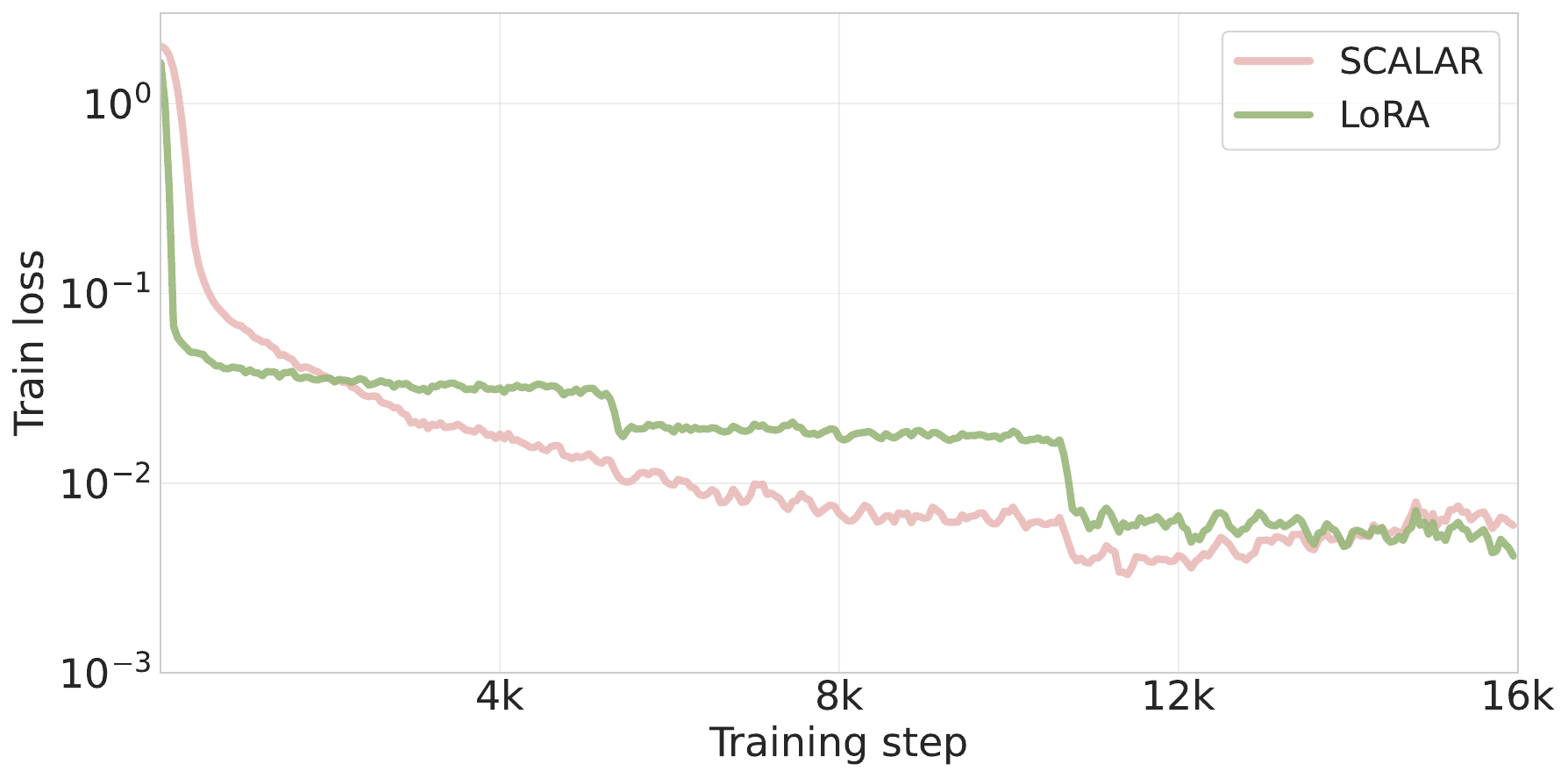}
        \caption{LLaMA3-8B}
    \end{subfigure}
    \caption{\textbf{Optimization behavior of rank-one LoRA and learned-direction
    reweighting.}
    The reweighting model freezes the directions learned by LoRA and
    re-optimizes one scalar coefficient for each adapted matrix. It optimizes
    $224$ coefficients for LLaMA3-8B and $196$ for Qwen2.5-7B, and follows a
    similar loss-reduction trajectory in these runs.}
    \label{fig:direction_reweighting_loss}
    \label{fig:oracle_scalar_lora}
\end{figure}

Figure~\ref{fig:direction_reweighting_loss} shows that the scalar-reweighted
model follows a similar training-loss trajectory to rank-one LoRA while
optimizing several orders of magnitude fewer parameters. Re-optimizing the
magnitudes of these learned directions recovers similar fitting behavior.

\begin{table}[!ht]
    \centering
    \caption{\textbf{Average accuracy of rank-one LoRA and learned-direction
    reweighting on eight commonsense reasoning tasks.}
    The reweighting model uses fewer than $0.01\%$ as many trainable parameters
    as rank-one LoRA and attains similar reported averages in these two
    settings.}
    \label{tab:direction_reweighting_accuracy}
    \label{tab:oracle_scalar_accuracy}
    \begin{tabular}{ll|lc}
        \toprule
        Model & Method & Params & Avg. Acc. \\
        \midrule
        \multirow{2}{*}{LLaMA3-8B}
        & LoRA ($r=1$) & 2,621,440 & 87.2 \\
        & Direction reweighting & 224 & 87.5 \\
        \midrule
        \multirow{2}{*}{Qwen2.5-7B}
        & LoRA ($r=1$) & 2,523,136 & 88.5 \\
        & Direction reweighting & 196 & 88.7 \\
        \bottomrule
    \end{tabular}
\end{table}

Table~\ref{tab:direction_reweighting_accuracy} reports comparable average
accuracy for direction reweighting on both model families. Together with
the training curves, these results show that re-optimizing the magnitudes
of learned rank-one LoRA directions recovers similar fitting behavior and
downstream accuracy. This motivates \name's construction of a calibrated
input basis before fine-tuning, followed by optimization of layer-specific
atom weights.

\subsection{Does LoRA Learn High-Energy Input Subspaces?}
\label{app:lora_subspace_energy}

We examine how the activation coverage of LoRA's input subspace changes
during fine-tuning. This diagnostic uses the coverage criterion defined in
Section~\ref{sec:motivation} to describe the learned subspaces.

We train LoRA adapters on a 7B language model. For an adapted linear
layer, let $\bm{A}^{t}\in\mathbb{R}^{R\times n}$ denote the input-side
LoRA factor at training step $t$. Directly measuring
$\|\bm{A}^{t}\bm{x}\|_2^2$ is not reliable because the LoRA
factorization is scale-indeterminate:
\begin{equation}
    \bm{B}^{t}\bm{A}^{t}
    =
    \left(c^{-1}\bm{B}^{t}\right)
    \left(c\bm{A}^{t}\right)
    \nonumber
\end{equation}
for any nonzero scalar $c$. We therefore remove the scale effect by
evaluating the row-normalized representative
\begin{equation}
    \overline{\bm{A}}^{t}
    =
    \left(
        \bm{A}^{t}(\bm{A}^{t})^{\top}
        +
        \epsilon\bm{I}_{R}
    \right)^{-\frac{1}{2}}
    \bm{A}^{t},
    \nonumber
\end{equation}
where $\epsilon$ is a small constant for numerical stability. Left
multiplication by the positive-definite factor preserves the row subspace of
$\bm{A}^{t}$. When $\epsilon=0$ and $\bm{A}^{t}$ has full row rank, the
rows of $\overline{\bm{A}}^{t}$ are orthonormal and the construction is
exactly invariant to positive global rescaling of $\bm{A}^{t}$. The small
regularizer used in computation makes this invariance approximate near
rank-deficient states.

During training, we reuse the input activations from the normal forward
pass and compute the captured activation-energy ratio
\begin{equation}
    \rho^{t}
    =
    \frac{
        \mathbb{E}_{\bm{x}}\left[
            \left\|\overline{\bm{A}}^{t}\bm{x}\right\|_2^2
        \right]
    }{
        \mathbb{E}_{\bm{x}}\left[
            \|\bm{x}\|_2^2
        \right]
    }.
    \label{eq:lora_energy_ratio}
\end{equation}
To compare different layers and module types, we report the relative
ratio
\begin{equation}
    \widehat{\rho}^{t}
    =
    \frac{\rho^{t}}{\rho^{0}},
    \nonumber
\end{equation}
where $\rho^{0}$ is the value at the random-initialized LoRA state.
Thus, $\widehat{\rho}^{t}>1$ indicates that the current LoRA input
subspace captures more normalized activation energy than its
initialization.

\begin{figure*}[!th]
    \centering
    \begin{subfigure}[b]{0.49\linewidth}
        \centering
        \includegraphics[width=\linewidth]{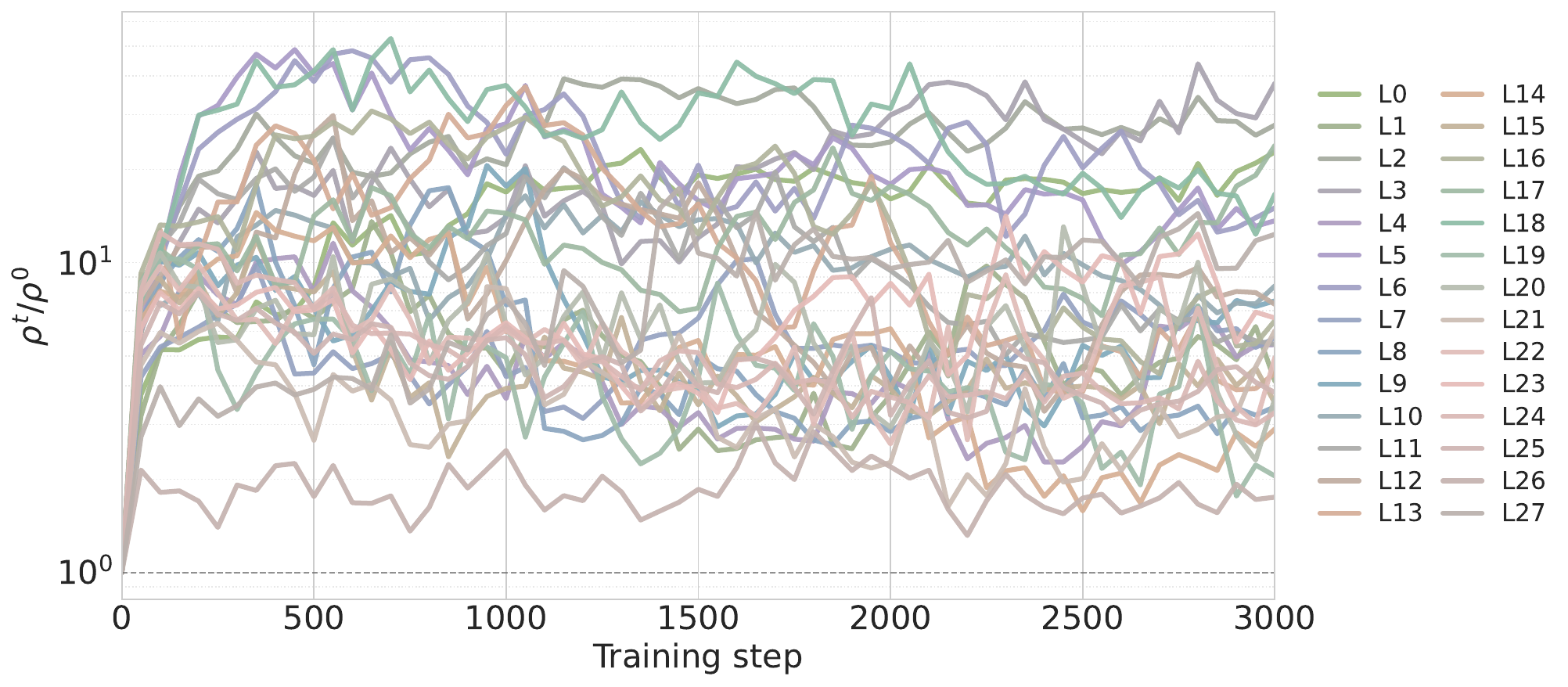}
        \caption{Query}
    \end{subfigure}
    \hfill
    \begin{subfigure}[b]{0.49\linewidth}
        \centering
        \includegraphics[width=\linewidth]{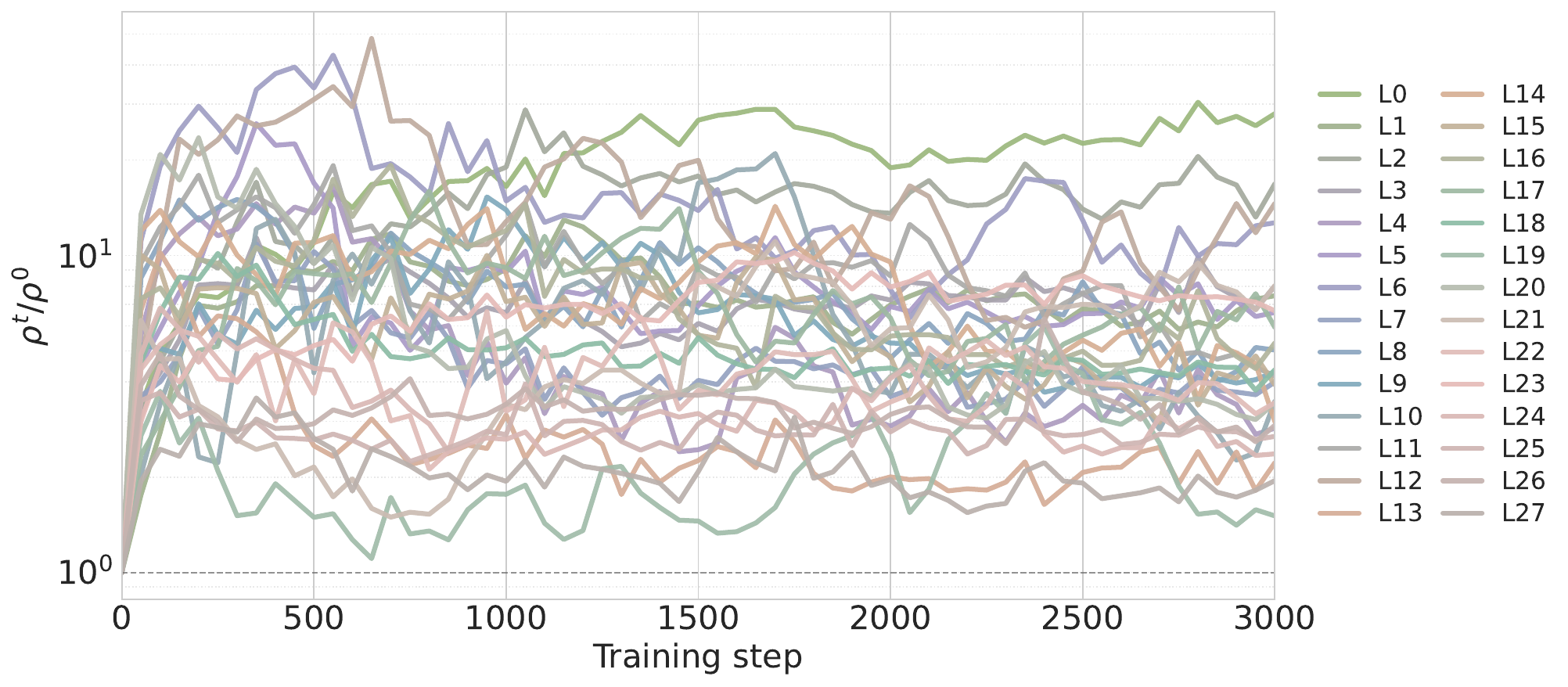}
        \caption{Key}
    \end{subfigure}
    \begin{subfigure}[b]{0.49\linewidth}
        \centering
        \includegraphics[width=\linewidth]{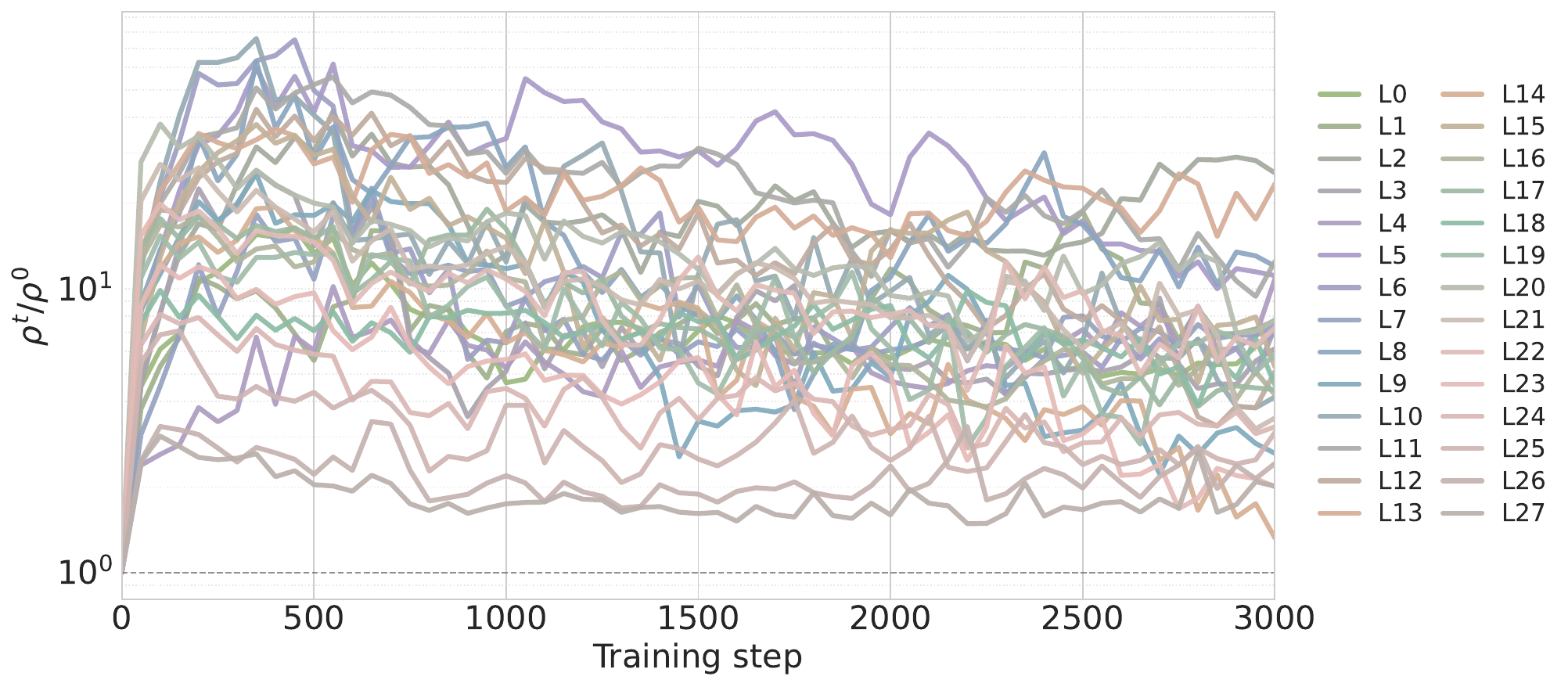}
        \caption{Value}
    \end{subfigure}
    \hfill
    \begin{subfigure}[b]{0.49\linewidth}
        \centering
        \includegraphics[width=\linewidth]{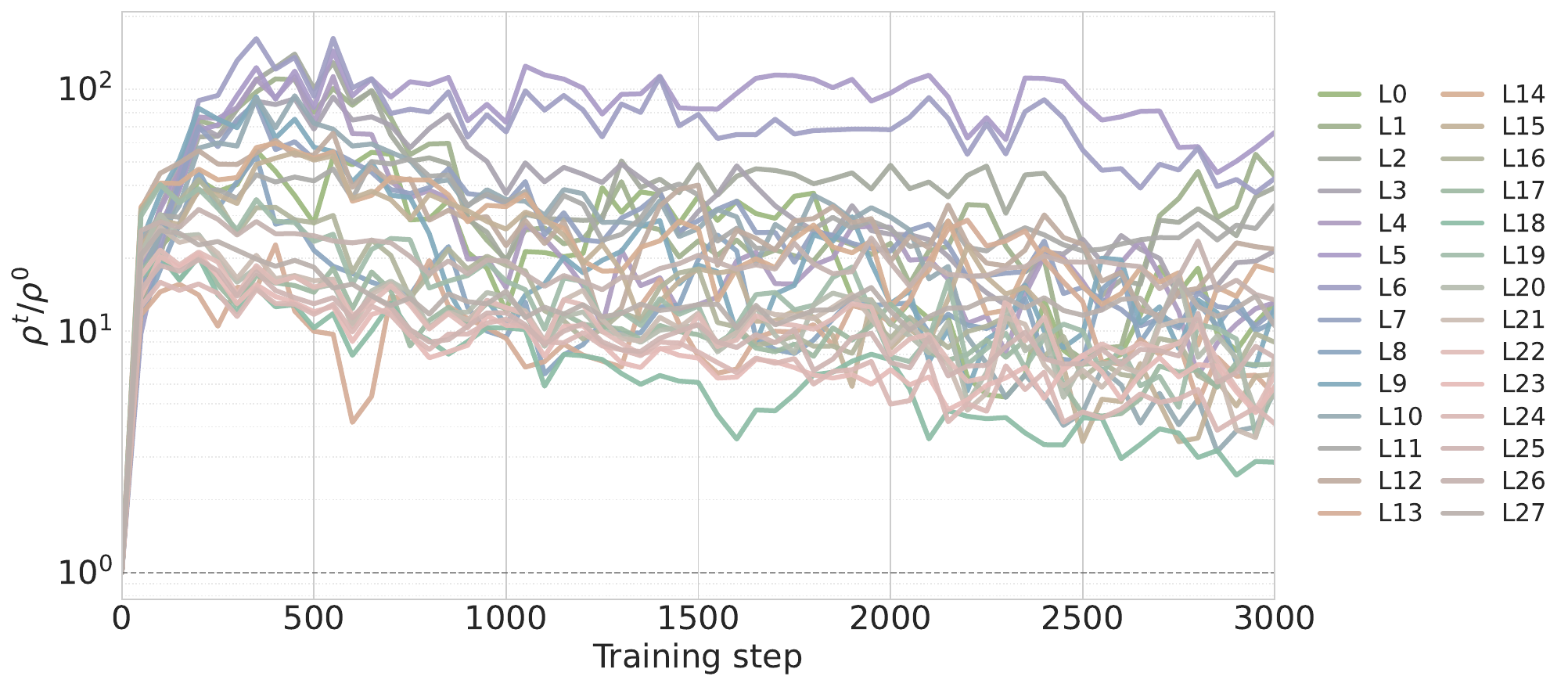}
        \caption{Output}
    \end{subfigure}
    \begin{subfigure}[b]{0.49\linewidth}
        \centering
        \includegraphics[width=\linewidth]{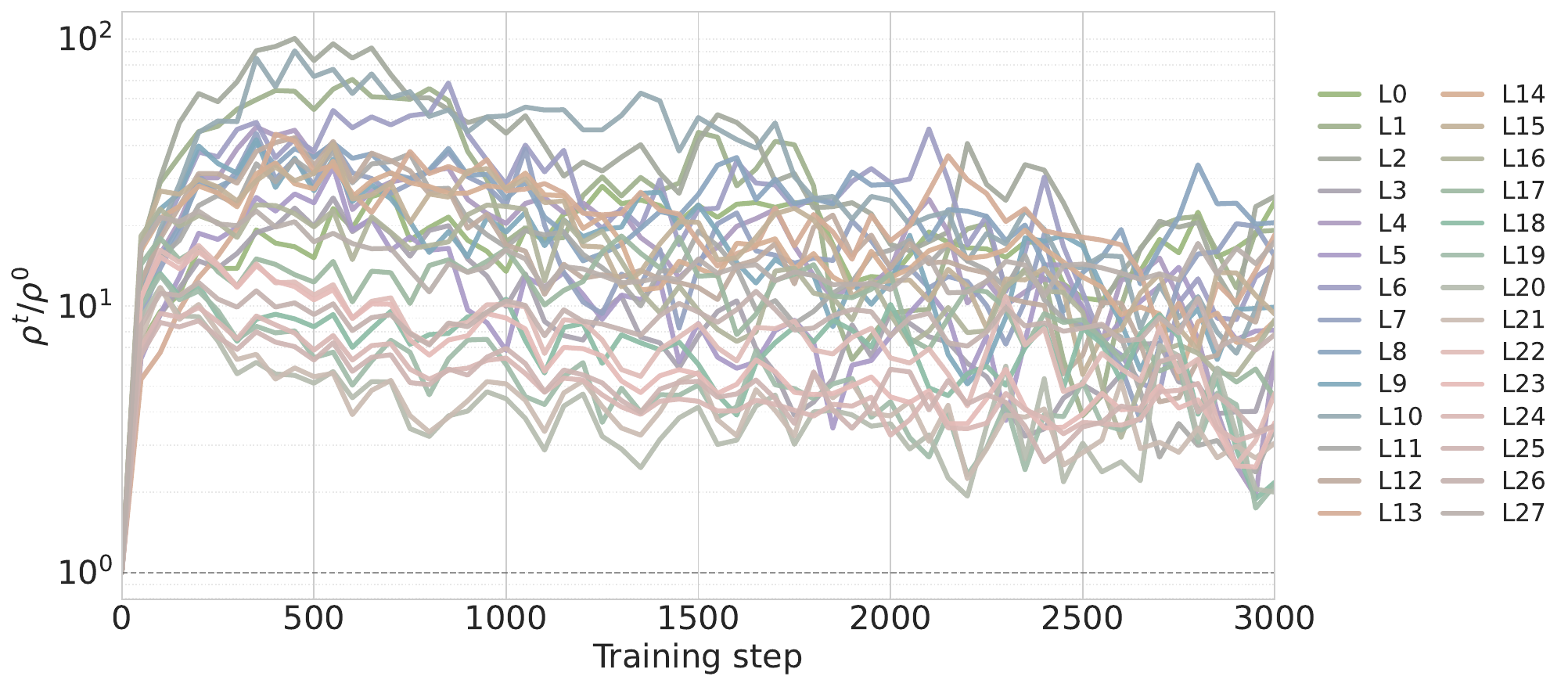}
        \caption{Up projection}
    \end{subfigure}
    \hfill
    \begin{subfigure}[b]{0.49\linewidth}
        \centering
        \includegraphics[width=\linewidth]{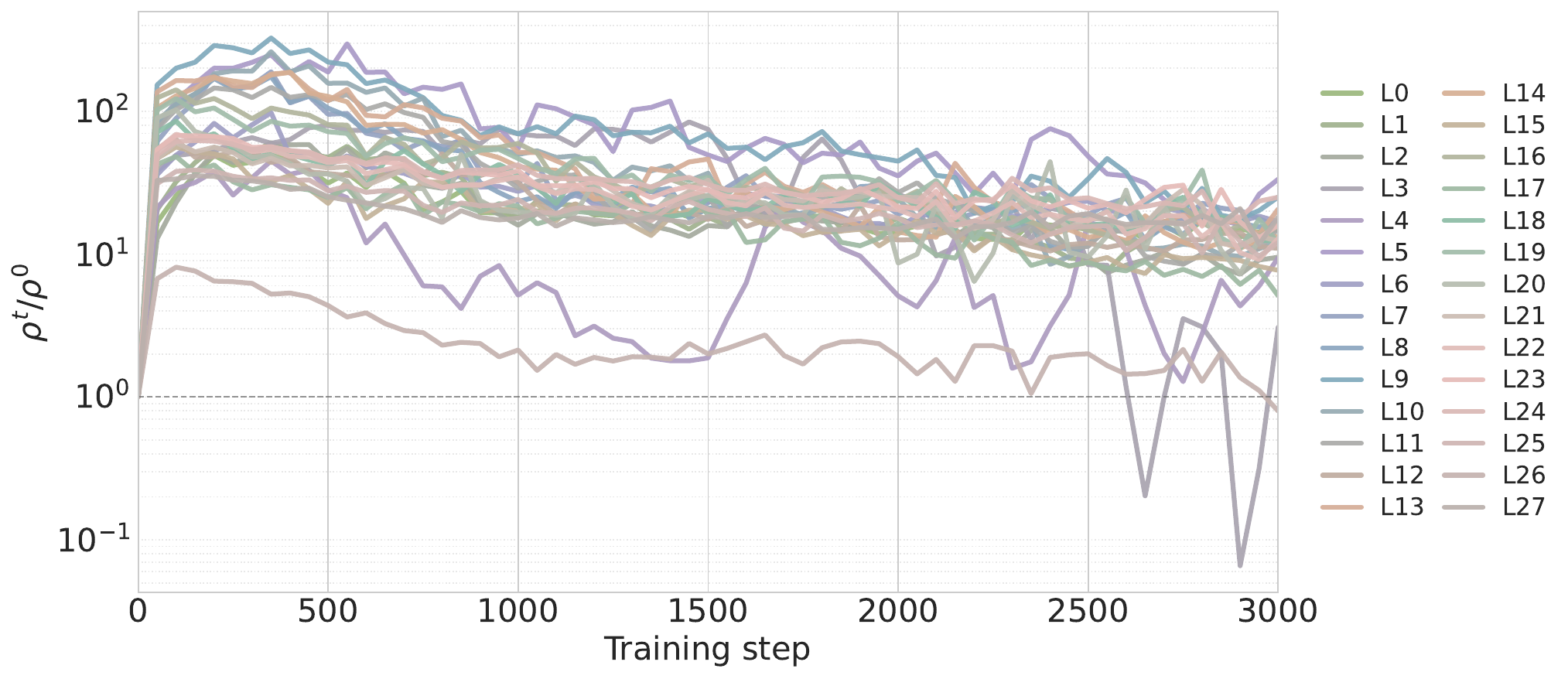}
        \caption{Down projection}
    \end{subfigure}
    \begin{subfigure}[b]{0.49\linewidth}
        \centering
        \includegraphics[width=\linewidth]{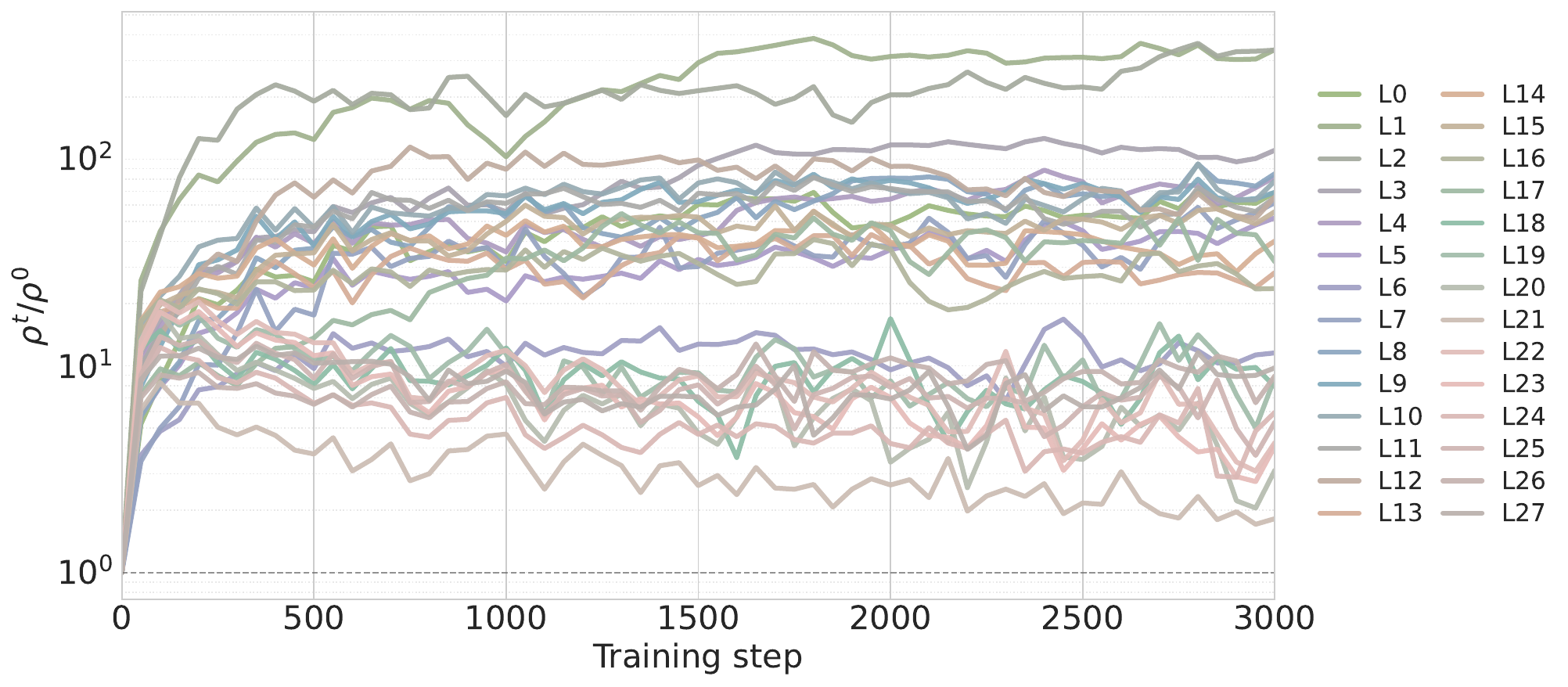}
        \caption{Gate projection}
    \end{subfigure}
    \caption{\textbf{Diagnostic training dynamics of LoRA on Qwen2.5-7B.} We report
    the relative normalized captured activation-energy ratio
    $\widehat{\rho}^{t}=\rho^{t}/\rho^{0}$, where $\rho^{t}$ is
    defined in \eqref{eq:lora_energy_ratio}. Values above $1$
    indicate stronger activation alignment than the
    random-initialized LoRA input subspace. Most module groups in this
    run show higher normalized activation coverage, with variation
    across layers and training steps.}
    \label{fig:lora_subspace_energy}
\end{figure*}

Figure~\ref{fig:lora_subspace_energy} reports
$\widehat{\rho}^{t}$ across adapted module types. The trajectories vary
across layers and training steps. Most module groups show higher normalized
activation coverage than at initialization, while one down-projection layer
falls below its initial value. These measurements describe the activation
coverage of subspaces learned by optimizing the task loss. They provide
empirical motivation for constructing an activation-calibrated input basis
before fine-tuning, as in \name.

\subsection{Comparison with Enhanced LoRA Variants}
\label{app:enhanced_lora}
\label{sec:enhanced_lora}

We further compare \name with DoRA~\cite{liu2024dora},
PiSSA~\cite{meng2024pissa}, and EVA~\cite{paischer2026parametereva}, which represent
three complementary improvements over standard LoRA. DoRA introduces a
magnitude--direction weight decomposition, PiSSA initializes and optimizes
the principal singular components of the pretrained weights, and EVA
initializes LoRA from high-variance activation directions with optional
adaptive rank allocation. These comparisons place \name alongside methods that improve the LoRA
parameterization or its initialization. EVA is particularly relevant because it also exploits
activation geometry; however, unlike EVA, which subsequently trains both
LoRA factors, \name freezes the activation-derived input basis and learns
only scalar combinations of fixed random output atoms.

All methods are evaluated on Qwen2.5-7B using the same training and
evaluation protocol. We use identical adapted modules and LoRA ranks for
LoRA, DoRA, PiSSA, and EVA, and independently tune their learning rates
using the same search budget.

\begin{table}[!ht]
    \centering
    \caption{\textbf{Comparison with enhanced LoRA variants on Qwen2.5-7B.}
    \name achieves comparable reported accuracy with fewer trainable
    parameters than the evaluated LoRA variants.}
    \label{tab:enhanced_lora_comparison}
    \begin{tabular}{l|c|c}
        \toprule
        \textbf{Method} & \textbf{Params} (\%) & \textbf{Avg. Acc.} \\
        \midrule
        LoRA & 0.0331 & 88.5 \\
        DoRA & 0.0513 & 88.7 \\
        PiSSA & 0.0331 & 88.5 \\
        EVA & 0.0331 & 88.7 \\
        \midrule
        \rowcolor{blue4}\name & 0.0105 & 88.9 \\
        \bottomrule
    \end{tabular}
\end{table}

Table~\ref{tab:enhanced_lora_comparison} reports average accuracies from
\(88.5\) to \(88.9\) across the evaluated methods. \name reaches
\(88.9\) using approximately one-third of the trainable parameters of
LoRA, PiSSA, and EVA, and about one-fifth of those used by DoRA. The
comparison supports competitive accuracy at a smaller trainable-parameter
budget.

EVA and \name both use activation-derived input directions. EVA then
optimizes the LoRA factors, while \name retains the calibrated input basis
and learns coefficients over its fixed random output dictionary. The
reported results compare the resulting parameter--performance trade-offs
under the stated configurations.

\subsection{General Instruction Fine-tuning}
\label{sec:general_instruction_tuning}

We evaluate post-tuning general-domain knowledge performance after
instruction fine-tuning on three model families spanning different
parameter scales: Gemma3-1B~\cite{gemma3},
LLaMA3.2-3B~\cite{grattafiori2024llama3}, and Qwen2.5-7B~\cite{qwen25report}. For each model, we fine-tune on the
52K-example Alpaca-GPT4 dataset~\cite{peng2023instruction} for 3 epochs and evaluate the resulting
models on MMLU~\cite{hendrycks2021mmlu} (5-shot). We report accuracy over the four MMLU subject categories,
including STEM, humanities, social sciences, and other domains, together
with the overall accuracy. Full-parameter fine-tuning, LoRA ($r=8$), and two
parameter configurations of \name use the same training data, number of
epochs, target modules, and optimization budget for a fair comparison.

\begin{table}[!ht]
    \centering
    \caption{\textbf{Post-tuning knowledge performance on MMLU.} Params denotes the percentage of
    trainable parameters. The best parameter-efficient result for each
    model is highlighted in bold.}
    \label{tab:general_instruction_tuning}
    \resizebox{\linewidth}{!}{
    \begin{tabular}{l|l|c|cccc|c}
        \toprule
        Model & Method & Params (\%) & STEM & Humanities &
        Social Sciences & Other & Overall \\
        \midrule
        & FT & 100.00 & 27.37 & 27.04 & 24.54 & 25.97 & 26.33 \\
        & LoRA & 0.6481 & 27.85 & 26.82 & 24.67 & 26.65 & 26.54 \\
        \rowcolor{blue4}\cellcolor{white} & \name ($K=1024$) & 0.0186 & 27.66 & 26.84 & 26.06 & 26.62 & 26.81 \\
        \rowcolor{blue3}\cellcolor{white}\multirow{-4}{*}{Gemma3-1B} & \name ($K=2048$) & 0.0373 & 27.78 & 27.10 & 25.67 & 27.97 & \textbf{27.13} \\
        \midrule
        & FT & 100.00 & 48.33 & 51.94 & 65.68 & 64.34 & 56.89 \\
        & LoRA & 0.3770 & 48.02 & 51.82 & 64.97 & 64.15 & 56.57 \\
        \rowcolor{blue4}\cellcolor{white} & \name ($K=2048$) & 0.0125 & 47.70 & 51.92 & 64.93 & 63.76 & 56.44 \\
        \rowcolor{blue3}\cellcolor{white}\multirow{-4}{*}{LLaMA3.2-3B} & \name ($K=4096$) & 0.0250 & 47.80 & 52.18 & 65.42 & 64.37 & \textbf{56.79} \\
        \midrule
        & FT & 100.00 & 71.77 & 68.06 & 84.24 & 77.21 & 74.46 \\
        & LoRA & 0.2643 & 71.39 & 68.33 & 84.11 & 77.05 & 74.41 \\
        \rowcolor{blue4}\cellcolor{white} & \name ($K=2048$) & 0.0052 & 71.55 & 68.12 & 84.69 & 77.41 & \textbf{74.58} \\
        \rowcolor{blue3}\cellcolor{white}\multirow{-4}{*}{Qwen2.5-7B} & \name ($K=4096$) & 0.0105 & 71.49 & 68.20 & 84.30 & 77.05 & 74.43 \\
        \bottomrule
    \end{tabular}
    }
\end{table}

Table~\ref{tab:general_instruction_tuning} shows comparable post-tuning
MMLU accuracy across the evaluated adaptation methods. The higher-scoring
\name configurations obtain 27.13 on Gemma3-1B, 56.79 on LLaMA3.2-3B,
and 74.58 on Qwen2.5-7B, using 0.0373\%, 0.0250\%, and 0.0052\%
trainable parameters, respectively.

The relative performance of the two \name configurations varies across
model families. The larger configuration has a higher reported average on
Gemma3-1B and LLaMA3.2-3B, while the smaller configuration has a higher
average on Qwen2.5-7B. Across these settings, \name achieves post-tuning
knowledge performance close to LoRA and full-parameter fine-tuning with a
smaller trainable-parameter budget.

\subsection{Scaling to Larger Models}
\label{app:large_model_scaling}

A key property of \name is that the number of trainable parameters
introduced for each adapted matrix is decoupled from its input and output
dimensions. For a weight matrix
$\bm{W}\in\mathbb{R}^{m\times n}$, a rank-$r$ LoRA adapter introduces
$r(m+n)$ trainable parameters. In contrast, \name optimizes only $K$
scalar coefficients for the corresponding fixed atoms, independent of
$m$ and $n$. Its total trainable parameter count may still depend on the
number of adapted modules and Transformer layers, but it does not grow
with the width of each linear transformation.

To examine whether this dimension-decoupled parameterization remains
effective on larger backbones, we extend the commonsense reasoning
experiments to Qwen2.5-14B and Qwen2.5-32B. All methods are trained on
Commonsense170K and evaluated on the same eight benchmarks used in the
main experiments. We use the same target modules and training protocol,
and compare two configurations of LoRA and \name.

\begin{table}[!ht]
    \centering
    \caption{\textbf{Commonsense reasoning results on larger Qwen2.5 models.}
    Params denotes the percentage of trainable parameters.
    LoRA introduces $r(m+n)$ trainable parameters for an
    $m\times n$ weight matrix, whereas \name assigns $K$
    trainable scalar coefficients to each adapted module,
    independent of $m$ and $n$.}
    \label{tab:large_model_scaling}
    \resizebox{\linewidth}{!}{%
    \begin{tabular}{l|l|c|cccccccc|c}
        \toprule
        Model & Method & Params (\%) & BoolQ & PIQA & SIQA & HS & WG & ARC-E & ARC-C & OBQA & Avg. \\
        \midrule
        \cellcolor{white} & LoRA ($r=1$) & 0.0291 & 78.6 & 93.1 & 84.9 & 97.4 & 92.3 & 97.7 & 93.4 & 95.8 & 91.7 \\
        \cellcolor{white} & LoRA ($r=8$) & 0.2324 & 77.8 & 93.7 & 83.8 & 97.6 & 92.3 & 97.9 & 93.4 & 96.8 & 91.7 \\
        \rowcolor{blue4}\cellcolor{white} & \name ($K=4096$) & 0.0093 & 77.8 & 94.0 & 84.9 & 97.4 & 91.9 & 97.9 & 93.8 & 96.6 & 91.8 \\
        \rowcolor{blue3}\cellcolor{white}\multirow{-4}{*}{Qwen2.5-14B} & \name ($K=8192$) & 0.0186 & 78.0 & 93.4 & 84.3 & 97.6 & 92.7 & 97.9 & 93.4 & 97.8 & \textbf{91.9} \\
        \midrule
        \cellcolor{white} & LoRA ($r=1$) & 0.0255 & 80.0 & 94.7 & 84.0 & 97.7 & 92.7 & 98.7 & 95.6 & 97.4 & 92.6 \\
        \cellcolor{white} & LoRA ($r=8$) & 0.2044 & 80.6 & 94.3 & 84.5 & 98.0 & 93.8 & 98.5 & 96.0 & 97.2 & \textbf{92.9} \\
        \rowcolor{blue4}\cellcolor{white} & \name ($K=4096$) & 0.0056 & 79.2 & 94.6 & 85.5 & 97.8 & 93.4 & 98.7 & 95.8 & 97.0 & 92.8 \\
        \rowcolor{blue3}\cellcolor{white}\multirow{-4}{*}{Qwen2.5-32B} & \name ($K=8192$) & 0.0112 & 79.7 & 94.5 & 85.7 & 97.7 & 94.1 & 98.6 & 95.5 & 97.2 & \textbf{92.9} \\
        \bottomrule
    \end{tabular}%
    }
\end{table}

As shown in Table~\ref{tab:large_model_scaling}, \name remains
competitive as the backbone size increases. On Qwen2.5-14B,
\name with $K=8192$ achieves the best average accuracy of $91.9$ while
training only $0.0186\%$ of the model parameters. The $K=4096$
configuration also slightly outperforms both LoRA settings, despite using
approximately $3.1\times$ fewer trainable parameters than rank-one LoRA.

On Qwen2.5-32B, \name with $K=8192$ matches rank-eight LoRA at an
average accuracy of $92.9$, while reducing the trainable parameter
percentage from $0.2044\%$ to $0.0112\%$, corresponding to an
approximately $18.3\times$ reduction. The $K=4096$ configuration remains
within $0.1$ points of the best result while training only $0.0056\%$ of
the model parameters.

For a fixed value of $K$, the trainable parameter percentage of \name
decreases as the backbone grows from 14B to 32B. This behavior follows
from the fact that the scalar budget of each adapted module is independent
of the matrix dimensions $m$ and $n$. By contrast, LoRA requires
$r(m+n)$ parameters per adapted matrix and therefore allocates more
trainable parameters as the model width increases. These results show
that \name preserves competitive adaptation performance on larger models
while becoming proportionally more parameter-efficient.

\subsection{Ablation Study}
\label{app:ablation}

We examine how grouping, cross-layer dictionary sharing, update rank,
calibration size, and output-atom initialization affect accuracy and
computational cost. The following subsections describe the corresponding
configurations.

\subsubsection{Grouped Parameterization and Cross-Layer Sharing}
\label{app:ablation_structure}

\noindent\textbf{Grouped parameterization.}
As introduced in Section~\ref{sec:method}, \name parameterizes the weight update as
\begin{equation}
    \Delta\bm{W}
    =
    s\sum_{r=1}^{R}\sum_{h=1}^{H}
    \gamma_{r,h}\bm{b}_{r,h}\bm{a}_{r}^{\top},
    \qquad
    K=RH,
    \nonumber
\end{equation}
where the \(H\) components in each group share the same input-side direction \(\bm{a}_{r}\), while retaining distinct output-side directions \(\bm{b}_{r,h}\). The update can therefore be equivalently expressed as
\begin{equation}
    \Delta\bm{W}
    =
    s\sum_{r=1}^{R}\widetilde{\bm{b}}_{r}\bm{a}_{r}^{\top},
    \qquad
    \widetilde{\bm{b}}_{r}
    =
    \sum_{h=1}^{H}\gamma_{r,h}\bm{b}_{r,h}.
    \label{eq:scalar-grouped-equivalent}
\end{equation}
Although the formulation contains \(K=RH\) scalar-weighted atoms, the \(H\) output atoms within each group can be merged into a single effective output vector \(\widetilde{\bm{b}}_{r}\) before being applied to the input activations. Consequently, the resulting update has rank at most \(R\), rather than \(K\).

This grouped structure also reduces the cost of applying the update. For a batch of activations \(\bm{X}\in\mathbb{R}^{n\times L}\), where \(L\) denotes the total number of tokens, the grouped update can be evaluated as
\begin{equation}
    \Delta\bm{W}\bm{X}
    =
    s\widetilde{\bm{B}}\bm{A}\bm{X},
    \nonumber
\end{equation}
where \(\bm{A}\in\mathbb{R}^{R\times n}\) and \(\widetilde{\bm{B}}\in\mathbb{R}^{m\times R}\). Constructing \(\widetilde{\bm{B}}\) requires weighted accumulation of the \(K\) fixed output atoms, with cost \(\mathcal{O}(mK)\). The subsequent matrix multiplications cost \(\mathcal{O}(LR(n+m))\), giving a total complexity of
\begin{equation}
    \mathcal{O}\bigl(mK+LR(n+m)\bigr).
    \nonumber
\end{equation}
Since \(K=RH\), the token-dependent matrix-multiplication cost scales with \(R\) rather than \(K\). The merging cost is independent of \(L\) and can be amortized over the batch and sequence dimensions. The grouped formulation also reduces the intermediate activation size from \(\mathcal{O}(LK)\) to \(\mathcal{O}(LR)\).

To examine the structural trade-off between the number of distinct input
directions and the per-token computation, we compare the grouped
parameterization with an ungrouped variant at the same coefficient budget:
\begin{equation}
    \Delta\bm{W}
    =
    s\sum_{k=1}^{K}\gamma_k\bm{b}_k\bm{a}_k^{\top},
    \label{eq:scalar-ungrouped-update}
\end{equation}
where each scalar coefficient is associated with an independent pair of input- and output-side directions. For \(\bm{A}\in\mathbb{R}^{K\times n}\) and \(\bm{B}\in\mathbb{R}^{m\times K}\), this update is evaluated as
\begin{equation}
    \Delta\bm{W}\bm{X}
    =
    s\bm{B}\operatorname{diag}(\bm{\gamma})\bm{A}\bm{X},
    \nonumber
\end{equation}
with dominant complexity
\begin{equation}
    \mathcal{O}\bigl(LK(n+m)\bigr).
    \nonumber
\end{equation}
The ungrouped variant can produce an update with rank as high as \(K\), but it requires all \(K\) input and output directions to be evaluated for every token. In contrast, grouping retains \(K\) candidate output atoms while reducing the token-dependent computation by approximately a factor of \(H\).

\begin{table}[!ht]
    \centering
    \caption{\textbf{Ablation on the grouped update parameterization.} Results are reported on the commonsense reasoning tasks using Qwen2.5-7B. The variants use the same coefficient budget and differ in the number of input directions, maximum update rank, and token-dependent computation.}
    \label{tab:ablation_group_ungroup}
    \begin{tabular}{l|ccc}
        \toprule
        Method & Iter./s & Mem. (GB) & Avg. Acc. \\
        \midrule
        Grouped, \eqref{eq:scalar-grouped-equivalent} \((K=1024,R=16)\)
        & 0.75 & 49.1 & 88.5 \\
        Ungrouped, \eqref{eq:scalar-ungrouped-update} \((K=1024)\)
        & 0.58 & 49.1 & 88.3 \\
        \bottomrule
    \end{tabular}
\end{table}

As shown in Table~\ref{tab:ablation_group_ungroup}, the grouped formulation increases training throughput from \(0.58\) to \(0.75\) iterations per second, corresponding to an improvement of approximately \(29\%\), without increasing peak memory. This result is consistent with the complexity analysis: the grouped implementation first merges the \(H\) output atoms associated with each input direction through weighted vector accumulation, and then performs matrix multiplications over only \(R\) effective directions. By contrast, the ungrouped variant evaluates \(K\) independent input and output directions for every token.

At this coefficient budget, the grouped and ungrouped variants obtain
average accuracies of \(88.5\) and \(88.3\), respectively. The grouped
variant combines comparable accuracy with higher throughput by allocating
\(K\) output atoms across \(R\) input directions. The coefficient
budget is unchanged, while the grouped computation applies fewer distinct
input and output projections to each token. This comparison shows the
system benefit of organizing the dictionary for efficient application.

\noindent\textbf{Cross-layer sharing of \(\bm{B}\).}
We next examine the cross-layer sharing strategy for the output-side bases.
In the default implementation, layers corresponding to the same projection
module share the same collection of fixed output directions
\(\{\bm{b}_{r,h}\}\). For example, all query-projection layers reuse one
output basis, while different module types retain separate bases.
Layer-specific adaptation is still preserved because each layer learns its
own coefficients \(\gamma_{r,h}\).

\begin{table}[!ht]
    \centering
    \caption{\textbf{Ablation on cross-layer sharing of \(\bm{B}\).}
    Sharing the output-side bases maintains nearly identical accuracy and
    throughput while substantially reducing memory consumption, especially
    as \(K\) increases.}
    \label{tab:ablation_cross_b}
    \begin{tabular}{l|ccc}
        \toprule
        Method & Iter./s & Mem. (GB) & Avg. Acc. \\
        \midrule
        Shared \(\bm{B}\) \((K=1024)\)
        & 0.75 & 49.1 & 88.5 \\
        Unshared \(\bm{B}\) \((K=1024)\)
        & 0.75 & 51.6 & 88.5 \\
        \midrule
        Shared \(\bm{B}\) \((K=2048)\)
        & 0.75 & 49.3 & 88.6 \\
        Unshared \(\bm{B}\) \((K=2048)\)
        & 0.75 & 54.2 & 88.6 \\
        \midrule
        Shared \(\bm{B}\) \((K=4096)\)
        & 0.74 & 49.7 & 88.9 \\
        Unshared \(\bm{B}\) \((K=4096)\)
        & 0.74 & 59.4 & 88.8 \\
        \bottomrule
    \end{tabular}
\end{table}

Table~\ref{tab:ablation_cross_b} reports nearly identical throughput and
average accuracy for the shared and unshared dictionaries at all three
values of \(K\). The memory benefit of sharing grows substantially
with \(K\). At \(K=1024\), sharing reduces peak memory from \(51.6\) GB to
\(49.1\) GB, while at \(K=4096\), the reduction increases from \(59.4\) GB
to \(49.7\) GB. Reusing the frozen output dictionaries reduces their
storage while retaining each layer's calibrated input basis and trainable
coefficients.

For the tested module dimensions and values of $K$, the shared and
unshared variants differ in average accuracy by at most $0.1$ point. Each
layer forms a distinct update through its calibrated input basis and scalar
coefficients. The two ablations connect grouping to higher training
throughput and dictionary sharing to lower peak memory at comparable
reported accuracy.

\subsubsection{Effect of the Update Rank}
\label{app:ablation_rank}

\begin{table}[!hb]
    \centering
    \caption{{\textbf{Effect of the update rank \(R\) under fixed trainable-parameter
    budgets.}}
    We report the average accuracy on the commonsense reasoning benchmarks
    using Qwen2.5-7B. Performance remains stable across a broad range of
    ranks, with \(R=16\) giving the strongest overall results.}
    \label{tab:ablation_rank}
    \begin{tabular}{l|ccc}
        \toprule
        & \(K=1024\) & \(K=2048\) & \(K=4096\) \\
        \midrule
        \(R=8\) & 88.4 & \textbf{88.6} & 88.7 \\
        \(R=16\) & \textbf{88.5} & \textbf{88.6} & \textbf{88.9} \\
        \(R=32\) & \textbf{88.5} & \textbf{88.6} & 88.8 \\
        \(R=64\) & 88.4 & 88.5 & 88.7 \\
        \bottomrule
    \end{tabular}
\end{table}

We investigate the effect of the maximum update rank \(R\) while keeping
the total number of trainable coefficients \(K=RH\) fixed. For each
\(K\in\{1024,2048,4096\}\), we vary
\(R\in\{8,16,32,64\}\) and set \(H=K/R\). Increasing \(R\) expands the
activation-derived input subspace and raises the maximum update rank,
whereas decreasing \(R\) allocates more output atoms to each input
direction.

As shown in Table~\ref{tab:ablation_rank}, \(R=16\) achieves the highest
or tied-highest reported accuracy under all three parameter budgets. The
variation across \(R=8\) to \(64\) is at most \(0.2\) points within
each value of \(K\). These results support \(R=16\) as a default
allocation between input-subspace coverage and the number of output atoms
per input direction.

\subsubsection{Effect of the Calibration Size}
\label{app:ablation_calibration_size}

\begin{table}[!ht]
    \centering
    \caption{\textbf{Effect of the calibration-set size on downstream performance
    and calibration cost on the 7B model.} Average accuracy is computed over the eight
    commonsense reasoning benchmarks. Calibration time denotes the one-time
    cost of collecting activation statistics and computing the input
    subspaces.}
    \label{tab:calibration_size}
    \begin{tabular}{l|cc}
        \toprule
        \textbf{Calibration size}
        & \textbf{Average accuracy}
        & \textbf{Calibration time (s)} \\
        \midrule
        16 & 88.3 & 73.483 \\
        32 & 88.5 & 108.18 \\
        64 & 88.4 & 147.67 \\
        128 (default) & 88.5 & 237.44 \\
        256 & 88.5 & 373.28 \\
        512 & 88.4 & 927.95 \\
        1024 & 88.5 & 1886.6 \\
        \bottomrule
    \end{tabular}
\end{table}

\name estimates the activation-derived input subspace from a calibration
set collected before fine-tuning. We therefore study how the number of
calibration examples affects both downstream performance and the one-time
cost of subspace construction. Table~\ref{tab:calibration_size} reports the
average accuracy over the eight commonsense reasoning benchmarks together
with the calibration time.

The downstream accuracy ranges from 88.3 to 88.5 across the evaluated
calibration sizes. Several sizes, including 32 and 128 examples, reach the
highest reported average of 88.5. The results show similar downstream
accuracy across a broad range of calibration-set sizes in this setting.

In contrast, the end-to-end calibration cost increases substantially with the
number of examples, from 73.4 seconds at a calibration size of 16 to
1886.6 seconds at 1024. This timing includes activation collection through
forward passes, distributed reduction of the activation Gram matrices, and
the subsequent SVD computation and basis application. We use 128 examples by
default, with an average accuracy of 88.5 and a calibration time of
237.44 seconds. This one-time calibration cost is reported separately from
the per-step training throughput in the main experiments.

\subsubsection{Effect of the Output-Atom Initialization}
\label{app:init_distribution}

In \name, the input basis \(\bm A\) is obtained from the leading right
singular vectors of downstream activations and remains fixed during
fine-tuning. The output atoms \(\bm B\) are randomly initialized and are
also kept frozen, while only the scalar coefficients are optimized. Under
the default Gaussian initialization, the entries of \(\bm B\) are sampled
independently as
\begin{equation}
    B_{ij}\sim\mathcal{N}(0,\sigma^2),
    \nonumber
\end{equation}
where \(\sigma\) denotes the element-wise standard deviation.

We examine whether performance is sensitive to the initialization
distribution or scale of the fixed output atoms. The upper block of
Table~\ref{tab:init_distribution} compares Gaussian, Rademacher, and
Kaiming initialization for \(\bm B\). The lower block retains Gaussian
initialization and varies \(\sigma\), while keeping the activation-derived
input basis and all other training hyperparameters unchanged.

\begin{table}[!ht]
    \centering
    \caption{\textbf{Effect of the initialization distribution and element-wise
    scale of the fixed output atoms \(\bm B\).} The upper block compares
    different initialization rules, while the lower block varies the
    Gaussian standard deviation \(\sigma\).}
    \label{tab:init_distribution}
    \begin{tabular}{l|cc}
        \toprule
        Output-atom initialization & \(K=1024\) & \(K=2048\) \\
        \midrule
        Gaussian (default) & 85.1 & \textbf{85.2} \\
        Rademacher & 85.1 & \textbf{85.2} \\
        Kaiming & 85.1 & 85.1 \\
        \midrule
        Gaussian, \(\sigma=0.05\) & 85.0 & \textbf{85.2} \\
        Gaussian, \(\sigma=0.10\) & 85.1 & \textbf{85.2} \\
        Gaussian, \(\sigma=0.20\) & 85.0 & 85.0 \\
        \bottomrule
    \end{tabular}
\end{table}

Gaussian, Rademacher, and Kaiming initialization yield reported accuracies
within \(0.1\) percentage points across the two values of \(K\).
Varying the Gaussian standard deviation from \(0.05\) to \(0.2\)
changes accuracy by at most \(0.2\) points. We use Gaussian
initialization with \(\sigma=0.1\) by default.

\section{Adapter Storage Analysis}
\label{app:adapter_storage}

Table~\ref{tab:param_comparison_baseline} compares trainable parameter
counts. Here we report the storage associated with the trainable
coefficients, frozen bases, and random dictionaries, together with the
representation used after merging. We distinguish trainable parameters,
compact adapter checkpoints, and deployable adapters. The checkpoint
comparison covers adapter tensors and basis-reconstruction metadata.

For a single adapted matrix
$\bm{W}\in\mathbb{R}^{m\times n}$, standard rank-$R$ LoRA stores
\begin{equation}
    \Delta\bm{W}=\bm{B}\bm{A},
    \qquad
    \bm{B}\in\mathbb{R}^{m\times R},
    \qquad
    \bm{A}\in\mathbb{R}^{R\times n}.
    \nonumber
\end{equation}
Thus, both its trainable parameter count and deployable adapter storage
are $R(m+n)$.

\name uses a different training-time parameterization:
\begin{equation}
    \Delta\bm{W}
    =
    s\sum_{r=1}^{R}\sum_{h=1}^{H}
    \gamma_{r,h}\bm{b}_{r,h}\bm{a}_{r}^{\top},
    \nonumber
\end{equation}
where $s$ is the fixed atom-scaling factor defined in
Appendix~\ref{app:atom_scaling}.
Only the $RH$ scalar coefficients are trainable. The input directions
$\bm{A}\in\mathbb{R}^{R\times n}$ are fixed task-specific buffers
obtained from activation statistics. During training, these input directions
require $Rn$ frozen values per adapted matrix. The output dictionary
contains $mRH$ frozen values per compatible sharing group, while the
coefficients and their optimizer states remain layer-specific. The random
output atoms can be regenerated from seeds for compact checkpoint storage.
After training, the output atoms associated with the same input
direction can be merged:
\begin{equation}
    \widetilde{\bm{b}}_{r}
    =
    \sum_{h=1}^{H}\gamma_{r,h}\bm{b}_{r,h}
    =
    \bm{B}_{r}\bm{\gamma}_{r}.
    \nonumber
\end{equation}
Therefore,
\begin{equation}
    \Delta\bm{W}
    =
    s\sum_{r=1}^{R}\widetilde{\bm{b}}_{r}\bm{a}_{r}^{\top}
    =
    s\widetilde{\bm{B}}\bm{A},
    \qquad
    \widetilde{\bm{B}}
    =
    \begin{bmatrix}
        \widetilde{\bm{b}}_{1} & \cdots & \widetilde{\bm{b}}_{R}
    \end{bmatrix}.
    \nonumber
\end{equation}
The deployable \name adapter is thus a standard rank-$R$ adapter and
requires $R(m+n)$ values, with the fixed scale $s$ absorbed into either
factor. During training, \name optimizes $RH$ coefficients per adapted
matrix. Its compact adapter checkpoint stores the calibrated $\bm{A}$,
the learned $\bm{\Gamma}$, and the seeds and sampling configuration used
to regenerate the random dictionary.

\begin{table*}[!ht]
    \centering
    \caption{Storage comparison for a single adapted matrix
    $\bm{W}\in\mathbb{R}^{m\times n}$. Here, $R$ is the adapter
    rank, $H$ is the number of \name output atoms per input
    direction, $K=RH$, and $N_{\mathrm{scale}}$ denotes the
    method-dependent number of stored scaling parameters for RandLoRA.
    Fixed random bases are assumed to be recoverable from seeds when
    applicable.}
    \label{tab:storage_symbolic}
    \setlength{\tabcolsep}{4.5pt}
    \renewcommand{\arraystretch}{1.18}
    \resizebox{\linewidth}{!}{
    \begin{tabular}{l|cccccc}
        \toprule
        Method
        & LoRA & LoRA-XS & VeRA & NOLA & RandLoRA & \cellcolor{blue4}\name \\
        \midrule
        Compact adapter checkpoint
        & $R(m+n)$
        & $R^2+R(m+n)$
        & $m+R+\mathrm{seeds}$
        & $k_A+k_B+\mathrm{seeds}$
        & $N_{\mathrm{scale}}+\mathrm{seeds}$
        & \cellcolor{blue4}$Rn+RH+\mathrm{seeds}$ \\
        \bottomrule
    \end{tabular}
    }
\end{table*}

For direct loading, the LoRA-XS checkpoint includes the trainable
$R\times R$ matrix and the frozen SVD-derived factors. Although these
frozen factors are not trainable, they are needed for direct adapter
loading unless one recomputes the truncated SVD from the pretrained
weight. For NOLA, random bases may be regenerated from seeds, so its compact
checkpoint need only retain the learned combination coefficients and the
corresponding seeds. \name has a similar
seed-based variant: instead of saving the random output dictionary, we
save only the activation-derived $\bm{A}$, the trained scalar matrix
$\bm{\Gamma}$, and the seeds required to regenerate
$\{\bm{B}_{r}\}_{r=1}^{R}$. When a deployable adapter is required,
$\bm{\Gamma}$ and the regenerated random dictionary are merged into
$\widetilde{\bm{B}}$, resulting in the standard rank-$R$ storage
$R(m+n)$.

We next instantiate these quantities on Qwen2.5-3B and Qwen2.5-7B. We
adapt all linear projection modules:
\texttt{q\_proj}, \texttt{k\_proj}, \texttt{v\_proj},
\texttt{o\_proj}, \texttt{gate\_proj}, \texttt{up\_proj}, and
\texttt{down\_proj}. Storage is reported under BF16 precision, where
each value occupies two bytes.

\begin{table*}[!t]
    \centering
    \caption{\textbf{Practical adapter storage on Qwen2.5-3B and Qwen2.5-7B under BF16 precision.} We report LoRA and \name with $R\in\{8,16\}$, while fixing $K=RH=2048$ for \name. Storage is reported in MB with $1\mathrm{MB}=10^6$ bytes.}
    \label{tab:storage_qwen}
    \resizebox{0.8\linewidth}{!}{
    \begin{tabular}{l|cc}
        \toprule
        Storage type & Qwen2.5-3B & Qwen2.5-7B \\
        \midrule
        Rank-$8$ LoRA deployable adapter & $29.93$ MB & $40.37$ MB \\
        \name compact checkpoint ($R=8$) & $14.45$ MB & $18.92$ MB \\
        \name deployable adapter ($R=8$) & $29.93$ MB & $40.37$ MB \\
        \midrule
        Rank-$16$ LoRA deployable adapter & $59.86$ MB & $80.74$ MB \\
        \name compact checkpoint ($R=16$) & $27.87$ MB & $37.04$ MB \\
        \name deployable adapter ($R=16$) & $59.86$ MB & $80.74$ MB \\
        \bottomrule
    \end{tabular}
    }
\end{table*}

On Qwen2.5-7B, the compact \name checkpoints occupy $18.92$ MB and
$37.04$ MB at ranks $8$ and $16$, respectively. The corresponding LoRA
adapters occupy $40.37$ MB and $80.74$ MB. Each compact \name checkpoint
stores the calibrated input basis and learned coefficients together with
dictionary-reconstruction metadata. The trainable scalar coefficients
account for $0.80$ MB in both cases because $K=2048$ is fixed; the
remaining tensor storage is the calibrated input basis. This compact
representation reduces the persistent storage required per adapter. After
materialization, both rank-$R$ adapters require $R(m+n)$ values and use
the standard low-rank execution form.

\section{Benchmark Details}
\label{app:benchmark_details}

We evaluate the adaptation methods on three representative domains:
commonsense reasoning, mathematical reasoning, and code generation. For each domain, we fine-tune the pretrained models on a corresponding
instruction dataset and evaluate the resulting checkpoints on held-out
benchmarks. The descriptions below summarize the data used in our experiments;
evaluation details and hyperparameters are given in
Appendix~\ref{app:hyperparameters}.

\subsection{Fine-Tuning Datasets}

\begin{itemize}

\item \textbf{Commonsense170K.}
Commonsense170K~\cite{hu2023llmadapters} is a collection of approximately
170K instruction-formatted examples constructed from the training sets of
eight commonsense reasoning benchmarks: BoolQ, PIQA, SIQA, HellaSwag,
WinoGrande, ARC-Easy, ARC-Challenge, and OpenBookQA. The original tasks
are converted into a unified instruction--response format and combined
for multi-task fine-tuning. Following prior PEFT studies, we train on the
combined dataset and evaluate separately on the test set of each
constituent benchmark.

\item \textbf{MetaMathQA.}
MetaMathQA~\cite{yu2023metamath} is a mathematical instruction-tuning
dataset containing approximately 395K examples. It is constructed by
rewriting existing mathematical questions from multiple perspectives,
including forward, backward, and answer-conditioned transformations,
while preserving the underlying mathematical knowledge. The dataset
contains step-by-step solutions and is designed to improve multi-step
mathematical reasoning. We fine-tune the models on MetaMathQA and evaluate
them on GSM8K and MATH.

\item \textbf{Code-Feedback.}
Code-Feedback~\cite{zheng2024codefeedback} contains approximately
66K code-related interactions involving code generation, execution
feedback, and iterative refinement. Its examples cover both initial
solutions and subsequent corrections based on execution results, providing
supervision for generating functionally correct programs. We use
Code-Feedback for code instruction tuning and evaluate the resulting
models on HumanEval and MBPP.

\end{itemize}

\subsection{Evaluation Benchmarks}

\noindent\textbf{Commonsense Reasoning.}
We evaluate commonsense reasoning using the eight benchmarks included in
Commonsense170K:

\begin{itemize}

\item \textbf{BoolQ}~\cite{clark2019boolq} consists of naturally occurring
yes-or-no questions paired with passages and evaluates reading
comprehension and implicit reasoning.

\item \textbf{PIQA}~\cite{bisk2020piqa} evaluates physical commonsense
reasoning by asking models to select the more plausible solution to an
everyday physical problem.

\item \textbf{SIQA}~\cite{sap2019socialqa} contains multiple-choice
questions about social interactions and evaluates reasoning about people's
intentions, reactions, and social consequences.

\item \textbf{HellaSwag}~\cite{zellers2019hellaswag} evaluates commonsense
natural-language inference by requiring models to select the most
plausible continuation of a given context.

\item \textbf{WinoGrande}~\cite{sakaguchi2021winogrande} is a large-scale
Winograd-style benchmark that tests commonsense reasoning and coreference
resolution through binary fill-in-the-blank questions.

\item \textbf{ARC-Easy and ARC-Challenge}~\cite{clark2018arc} are the easy
and challenging subsets of the AI2 Reasoning Challenge, respectively.
Both contain grade-school-level multiple-choice science questions, with
ARC-Challenge including questions that are more difficult for retrieval-
and word-matching-based systems.

\item \textbf{OpenBookQA}~\cite{mihaylov2018openbookqa} contains
multiple-choice science questions that require combining elementary
scientific facts with broader commonsense knowledge.

\end{itemize}

We report accuracy on each benchmark and use their unweighted mean as the
overall commonsense reasoning score.

\noindent\textbf{Mathematical Reasoning.}
We evaluate mathematical reasoning on GSM8K and MATH:

\begin{itemize}

\item \textbf{GSM8K}~\cite{cobbe2021gsm8k} contains 8.5K linguistically
diverse grade-school mathematics word problems. Solving these problems
generally requires several arithmetic reasoning steps before producing a
final numerical answer.

\item \textbf{MATH}~\cite{hendrycks2021math} consists of competition-level
mathematics problems spanning algebra, counting and probability,
geometry, intermediate algebra, number theory, pre-algebra, and
pre-calculus. Compared with GSM8K, it requires more advanced mathematical
knowledge and longer reasoning chains.

\end{itemize}

We extract the final answer from each model response and report exact-match
accuracy following the standard evaluation protocol.

\noindent\textbf{Code Generation.}
We evaluate code generation on HumanEval and MBPP:

\begin{itemize}

\item \textbf{HumanEval}~\cite{chen2021humaneval} contains 164
hand-written Python programming problems. Each problem provides a function
signature, a natural-language specification, and hidden unit tests used to
assess functional correctness.

\item \textbf{MBPP}~\cite{austin2021mbpp} contains short Python programming
problems described in natural language, together with reference solutions
and test cases. The tasks primarily assess the ability to synthesize
functionally correct programs for basic programming problems.

\end{itemize}

For both benchmarks, generated programs are executed against the provided unit
tests, and we report pass@1 as the primary metric.

\noindent\textbf{General-Domain Knowledge Evaluation.}
We evaluate post-tuning general-domain knowledge performance using
Massive Multitask Language Understanding
(MMLU)~\cite{hendrycks2021mmlu}. MMLU consists of multiple-choice
questions from 57 academic and professional subjects, covering STEM,
humanities, social sciences, and other domains. The benchmark evaluates
the breadth of a model's pretrained knowledge and its ability to apply
that knowledge across heterogeneous subject areas.

We report accuracy for the four aggregated subject categories and the
overall accuracy across all MMLU tasks. These scores characterize
post-tuning knowledge performance after fine-tuning on
Alpaca-GPT4~\cite{peng2023instruction}.

\end{document}